\documentclass{article}

\usepackage{arxiv}

\usepackage[utf8]{inputenc} % allow utf-8 input
\usepackage[T1]{fontenc}    % use 8-bit T1 fonts
\usepackage{hyperref}       % hyperlinks
\usepackage{url}            % simple URL typesetting
\usepackage{booktabs}       % professional-quality tables
\usepackage{amsfonts}       % blackboard math symbols
\usepackage{nicefrac}       % compact symbols for 1/2, etc.
\usepackage{microtype}      % microtypography
\usepackage{lipsum}		% Can be removed after putting your text content
\usepackage{graphicx}
\usepackage[numbers,sort&compress]{natbib}
\usepackage{doi}

\usepackage{multirow}
\usepackage{tabularx}
\usepackage{amsmath} 
\usepackage{subcaption}
\usepackage{tabularx}
\usepackage{longtable}
\usepackage{array}
\usepackage{booktabs} % optional but looks better

\title{A Survey of Freshness-Aware Wireless Networking with Reinforcement Learning}

\author{
Alimu Alibotaiken$^{*}$,
Suyang Wang$^{\dagger}$,
Oluwaseun T. Ajayi$^{*}$,
Mushu Li$^{\ddagger}$,
and Yu Cheng$^{*}$ \\[1ex]
$^{*}$Department of Electrical and Computer Engineering, Illinois Institute of Technology, Chicago, IL 60616, USA\\
$^{\dagger}$School of Computer Science and Engineering, California State University, San Bernardino, CA 92407, USA\\
$^{\ddagger}$Department of Computer Science and Engineering, Lehigh University, Bethlehem, PA 18015, USA\\[1ex]
Email: \{aalibotaiken, oajayi6\}@hawk.illinoistech.edu,
suyang.wang@csusb.edu,
mul224@lehigh.edu,
cheng@illinoistech.edu
}
\date{}

\renewcommand{\shorttitle}{A Survey of Freshness-Aware Wireless Networking with Reinforcement Learning}

\begin{document}
\maketitle

\begin{abstract}
The age of information (AoI) has become a central measure of data freshness in modern wireless systems. Reinforcement learning (RL) has emerged as a promising approach to AoI optimization, enabling adaptive decision-making in dynamic and stochastic wireless environments without requiring precise system models. Yet existing surveys either focus on classical AoI formulations or provide broad discussions of RL in wireless networks without addressing freshness as a unified learning problem. Motivated by this gap, this survey examines RL specifically through the lens of AoI and generalized freshness optimization. Motivated by this gap, this survey examines RL specifically through the lens of AoI and generalized freshness optimization. We organize AoI and its variants into native, function-based, and application-oriented families, providing a clearer view of how freshness should be modeled in B5G and 6G systems. Building on this foundation, we introduce a policy-centric taxonomy that reflects the decisions most relevant to freshness, consisting of update-control RL, medium-access RL, risk-sensitive RL, and multi-agent RL. This structure provides a coherent framework for understanding how learning can support important wireless networking problems, including sampling, scheduling, UAV trajectory planning, medium access, and distributed coordination. We further synthesize recent progress in RL-driven freshness control and highlight open challenges. The goal is to establish a unified foundation for learning-based freshness optimization in next-generation wireless networks.
\end{abstract}

% keywords can be removed

\section{Introduction}
% The very first letter is a 2 line initial drop letter followed
% by the rest of the first word in caps.
% 
% form to use if the first word consists of a single letter:
% \IEEEPARstart{A}{demo} file is ....
% 
% form to use if you need the single drop letter followed by
% normal text (unknown if ever used by the IEEE):
% \IEEEPARstart{A}{}demo file is ....
% 
% Some journals put the first two words in caps:
% \IEEEPARstart{T}{his demo} file is ....
% 
% Here we have the typical use of a "T" for an initial drop letter
% and "HIS" in caps to complete the first word.
In the rapidly evolving landscape of wireless communication, the efficient management of information flow has become increasingly important. With the growing presence of Internet of Things (IoT) devices, autonomous systems, and real-time applications, networks must deliver information in a timely and dependable manner. These developments have brought renewed attention to the notion of information freshness, which traditional metrics such as delay or throughput do not fully capture. To describe freshness more precisely, the age of information (AoI) \cite{6195689} has emerged as a key performance metric, measuring how up-to-date the data at the destination is.

AoI is defined as the time elapsed since the generation of the most recently received update, characterizing the timeliness of status updates in status updating systems and applications \cite{9380899}. AoI optimization is not equivalent to minimizing delay or maximizing the source sampling rate. For instance, a low sampling rate can avoid unnecessary queueing delay, while the timeliness of the information at the receiver is unfavorable due to infrequent updates. However, too frequent updates would increase queueing delay due to congestion, thereby causing AoI degradation \cite{7524524,7283009}. As a result, AoI optimization has attracted significant interest in various wireless network settings, including sensor networks, vehicular networks, and mobile edge computing.

%Some of the early studies on AoI minimization explored several traditional approaches, such as convex optimization, queuing theory, Lyapunov optimization, and heuristic scheduling. For networks with general interference constraints, AoI could be formulated as a convex optimization problem \cite{8943134}, solvable by algorithms like Frank-Wolfe algorithm \cite{article}. Convex optimization has also been used to design efficient channel access and scheduling, which ensures optimal trade-offs between AoI and network constraints \cite{9525063,8445979,10620720,9023385}. Queuing theory uses tools like renewal theory and generating functions \cite{9377627} to derive the analytical expressions for AoI and its stationary distribution. Such a framework helps to evaluate and optimize different scheduling policies (e.g. first come first serve (FCFS), last come first serve (LCFS), preemptive decisions). Lyapunov optimization techniques transform AoI minimization into real-time control problems, enabling adaptive scheduling in multi-user and multi-hop networks \cite{8476220},\cite{9484621}. Lastly, heuristic methods (e.g. maximum age first (MAF), randomized scheduling, Whittle's index policy) provide computationally efficient strategies \cite{8514816}.  

Early studies on AoI minimization explored traditional analytical approaches, including convex optimization, queuing theory, Lyapunov optimization, and stochastic hybrid systems (SHS). For networks with general interference constraints, AoI can be cast as a convex optimization problem \cite{8943134} and solved using methods such as the Frank–Wolfe algorithm \cite{article}, while similar optimization techniques have been used to design channel access and scheduling policies that balance AoI with network constraints \cite{9525063,8445979,10620720,9023385}. Queuing-theoretic tools such as renewal theory and generating functions \cite{9377627} provide closed-form AoI expressions for various service disciplines \cite{6195689,8469047,6284003,8000687,7283009,8006543,10000608}, and Lyapunov optimization enables adaptive scheduling in multi-user and multi-hop systems \cite{8476220,9484621}. SHS offers another analytical viewpoint by modeling the joint discrete–continuous evolution of age, with applications ranging from finite-state queueing systems \cite{8469047} to Carrier-sense Multiple Access (CSMA)-based random access networks \cite{9007478} and \cite{10621330}, as well as recent learning-augmented SHS models for heterogeneous unsaturated CSMA settings \cite{11044711}. Lastly, heuristic strategies such as maximum-age-first scheduling, randomized access, and Whittle index-based policies offer lightweight and computationally efficient solutions \cite{8514816}.

%Notable methods include the zero-wait policy, threshold policy \cite{8647281}, and preemptive policy \cite{9155304}. The four low-complexity scheduling policies, maximum age first (MAF), stationary randomized, max-weight, and Whittle's Index, were considered in \cite{8514816}. For networks with general interference constraints, AoI could be formulated as a convex optimization problem \cite{8943134}, solvable by algorithms like Frank-Wolfe algorithm \cite{article}. 

Despite the promising results of these optimization techniques, these optimization techniques often require detailed and precise mathematical models of the network, which are typically unavailable for complex and dynamic systems. For instance, convex and queueing-based methods typically require accurate characterizations of arrival and service processes. Meanwhile, Lyapunov optimization depends on knowing exactly how the system evolves over time. If AoI is updated through random (due to stochastic network delays) or complex processing, the evolution may not be expressible in a closed-form. In that case, the model-driven AoI optimization methods cannot be applied directly because the system’s evolution is not clearly specified. Heuristic methods, while effective in specific scenarios, need to be manually redesigned for new problems, limiting their generalizability and also lack performance guarantees. Additionally, the AoI minimization in link transmission scheduling problem has been proven to be NP-hard \cite{8022894}, meaning that traditional optimization techniques struggle with the computational complexity, which can grow exponentially with the size of the network. Given these challenges and limitations, data-driven machine learning (ML) techniques have been explored to minimize AoI in various network settings \cite{shisher2021age, shisher2022does, leng2021learning, ndiaye2023ensemble, akbari2023constrained, wu2023towards, liu2021age, wang2022deep}. 

The success of ML in real-time decision-making applications (e.g., vehicle trajectory prediction \cite{10510276,aradi2020survey}, smart agriculture \cite{10690239,mahmood2023machine}, remote surgery \cite{9519658,9950359}, etc.), where maintaining fresh status updates is crucial, further demonstrate ML as a valuable approach for solving the AoI minimization problem. Specifically, reinforcement learning (RL) is a promising approach for AoI optimization due to its ability to adapt to complex, dynamic environments without requiring precise models in advance for decision making. RL is a branch of ML where an agent learns to make decisions by interacting with its environment to maximize cumulative rewards. Unlike traditional optimization techniques, RL does not require explicit mathematical models of the environment. Instead, it learns optimal policies through trial and error, adapting to changing conditions and complex dynamics. 

The primary advantage of RL is its ability to handle large-scale and high-dimensional problems, making it well-suited for optimizing AoI in wireless networks. RL algorithms can efficiently explore and exploit the environment to discover effective strategies for minimizing AoI, even in the presence of stochastic network delays and varying traffic loads. The adaptability and scalability of RL make it a powerful tool for enhancing information freshness in dynamic, heterogeneous network environments. Moreover, RL has demonstrated significant potential for solving NP-hard problems across various domains \cite{9785600}, \cite{DBLP:journals/corr/abs-2003-03600}. In comparison with traditional approaches for combinatorial optimization problems which often rely on handcrafted heuristics, and lead to suboptimal objective values, RL automates the search for these heuristics by training an agent in a supervised or self-supervised manner, providing a promising alternative \cite{DBLP:journals/corr/abs-2003-03600}.

\begin{table*}[t]
\centering
\caption{Comparison of Existing Information Freshness-related Surveys}
\footnotesize
\renewcommand{\arraystretch}{1.35}
\begin{tabularx}{\textwidth}{|>{\centering\arraybackslash}m{1cm}|X|m{2.5cm}|X|m{2.7cm}|}
\hline
\textbf{Survey} & 
\textbf{Freshness Metrics Discussed} &
\textbf{Machine Learning Methods} &
\textbf{Reinforcement Learning \& Policy-Centric Discussion} &
\textbf{Application Scenario / Scope} \\
\hline

\cite{9380899}  
& AoI, Peak AoI, average AoI, age-penalty functions (listed individually; no classification). 
& ML mentioned as future direction 
& None; RL papers only briefly cited. 
& General AoI theory across wireless networks \\
\hline

\cite{8930830}  
& AoI, PAoI, Value of Information Update (VoIU), age penalty function (briefly mentioned). 
& None 
& None 
& RF-powered IoT networks \\
\hline

\cite{9903385}  
& AoI, AoI-related penalty functions
& ML mentioned briefly for generic optimization and prediction 
& None
& Cellular IoT \\
\hline

\cite{ABBAS2023199}  
& AoI, PAoI, VoI  
& None 
& None
& Massive IoT / Ambient Intelligence networks. \\
\hline

\cite{10286022}  
& AoI, PAoI, AoX family (AUD, AoCI, etc.) 
& RL  
& RL papers listed, but no RL policy taxonomy 
& IoT systems \\
\hline

\cite{s23198238}  
& AoI, PAoI
& None 
& None 
& Cellular wireless networks (5G) \\
\hline

\cite{10893697}  
& AoI, Version AoI (vAoI)
& None  
& None 
& Gossip / information-spreading networks. \\
\hline

\cite{drones7040260}  
& AoI, PAoI (no classification)
& RL 
& RL (mentioned in referenced papers, not a focus)
& UAV-assisted WSN/IoT networks \\
\hline

\textbf{Our Survey}  
& Unified 3-class taxonomy: Native AoI (AoI, PAoI); Function-based AoI (penalties, statistics, violation metrics); Semantic/task-aware metrics (AoII, VoI, semantic freshness). 
& Systematic ML coverage: RL, MARL 
& First policy-centric RL synthesis: sampling, scheduling, routing, trajectory, and MAC access policies.
& Cross-domain: IoT, UAV, vehicular, cellular, edge/semantic networks \\
\hline

\end{tabularx}
\label{tab:aoi_survey_comparison}
\vspace{-1.8em}
\end{table*}

While several surveys have examined AoI and its role in wireless networks \cite{9380899,8930830,9903385,ABBAS2023199,s23198238,10893697,drones7040260}, none provide a RL–centered treatment of AoI optimization. Existing work primarily focuses on classical queueing analysis, traditional scheduling policies, or domain-specific architectures such as IoT, UAVs, and cellular systems, and thus does not explore how RL can systematically learn sampling, scheduling, trajectory, or access policies to minimize AoI. Moreover, prior surveys list AoI metrics individually, without providing a unified framework for understanding the expanding family of freshness measures. In contrast, this survey introduces a three-class taxonomy of AoI and its variants and presents the first policy-centric organization of RL methods for AoI, covering task-oriented, access-oriented, risk-sensitive, and multi-agent settings across diverse network models. This perspective establishes a comprehensive and coherent foundation for understanding how RL can be leveraged to optimize information freshness in modern wireless systems.

%The comparison of related survey papers is summarized in Table~\ref{tab:survey_comparison_extended}. 
%Each column highlights a key dimension of analysis. The \textit{Focus Area} identifies the primary topic or domain covered by each survey, while the \textit{AoI}, \textit{RL}, \textit{UAV}, \textit{MAC}, and \textit{MARL} columns capture whether the work addresses age of information, RL techniques, UAV-related applications, medium access control, or multi-agent RL (MARL), respectively. Beyond these baseline dimensions, we introduce three additional columns to reflect the unique contributions of our survey: \textit{Policy Taxonomy}, which denotes whether a unifying, policy-level categorization of RL approaches is provided; \textit{Freshness+}, which assesses whether the survey extends the discussion beyond classical AoI to function-aware, content-aware, or semantic freshness metrics; and \textit{X-Layer}, which captures whether cross-layer, end-to-end RL perspectives (spanning sampling, scheduling, MAC, and trajectory) are addressed. As shown in the table, existing surveys cover subsets of these aspects, but none provide an integrated framework that connects them. Our work uniquely addresses all dimensions, establishing a comprehensive and structured foundation for RL-driven AoI and freshness optimization in wireless networks. In addition, Fig.~\ref{fig:survey-structure} illustrates the overall structure of this survey.

The comparison of related survey papers is presented in Table~\ref{tab:aoi_survey_comparison}. Each column highlights a central aspect of how information freshness has been studied. The column on freshness metrics shows the scope of AoI variants discussed in each survey, while the column on machine learning indicates whether learning-based methods are incorporated in any meaningful way. The RL and policy-centric discussion column assesses whether RL is examined beyond a brief mention and whether the survey provides a structured view of how different RL methods relate to AoI optimization. Finally, the application scenario column summarizes the primary network domains considered by each work.

As the table shows, existing surveys tend to address only individual components of this landscape. Most focus on classical AoI measures without offering a unified view of metric variants. Many mention ML only briefly, and very few provide any organized treatment of RL. None introduces a policy-level framework that explains how RL methods map to different AoI optimization tasks across wireless systems.

To bridge this gap, this paper provides a systematic and forward-looking survey of how RL can be used to optimize AoI and generalized information freshness in wireless networks. Our key contributions include:
\begin{figure}[p]
    \centering
    \includegraphics[width=0.9\textwidth]{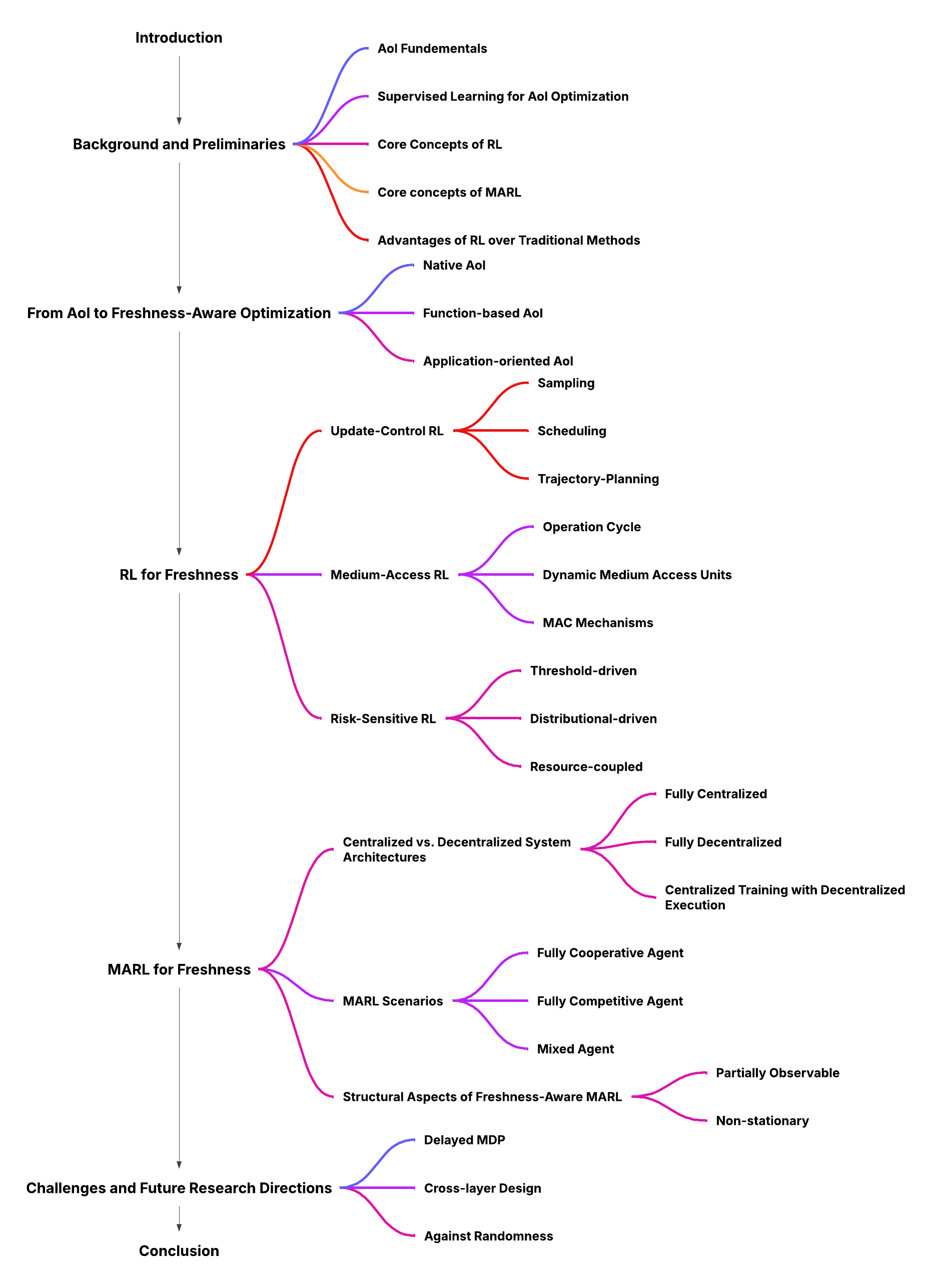}
    \caption{Overall structure of this survey.}
    \label{fig:survey-structure}
\end{figure}
\begin{itemize}
    \item \textbf{A modern reframing of AoI:} We organize AoI into three categories: native AoI, function-based AoI, and application-oriented freshness, showing how the idea of timeliness has evolved from simple recency toward formulations that account for resource use, risk, and task relevance.

    \item \textbf{A novel categorization framework:} We propose a new way to classify RL methods based on the policies they learn:
    \begin{itemize}
        \item \textbf{Update-Control RL} learning when and what to update through sampling, scheduling, and trajectory decisions.
        \item \textbf{Medium-Access RL} learning how devices contend for shared wireless resources at the MAC layer.
        \item \textbf{Risk-Sensitive RL} managing tail risks and stochastic variability.
    \end{itemize}

    \item \textbf{Multi agent reinforcement learning (MARL) for freshness:} We provide a systematic review about coordinating freshness-aware decisions among distributed agents.

    \item \textbf{Systematic review of RL for freshness in B5G/6G networks:} We synthesize recent RL methods for optimizing information freshness across key B5G/6G networking contexts, including ultra-dense wireless access, edge intelligence, semantic communications, and autonomous systems.

    \item \textbf{Future Directions:} We identify unresolved issues such as delayed MDPs, the need for cross-layer design and robustness against randomness, providing directions for future research in this field.
\end{itemize}

The overall structure of this survey is illustrated in Fig.~\ref{fig:survey-structure}. Section~\ref{background} provides background and preliminaries, including AoI fundamentals, supervised learning approaches, and key RL concepts. Section~\ref{freshness metrics} introduces a unified classification of freshness-aware metrics into native, function-based, and application-oriented formulations. Section~\ref{policy-centric} presents a policy-centric taxonomy of RL methods for freshness optimization, covering update-control, medium-access, and risk-sensitive policies. Section~\ref{MARL} discusses MARL for freshness optimization, including system architectures, agent interactions, and structural challenges. Section~\ref{future} identifies open challenges and future research directions, including delayed decision processes, cross-layer design, and robustness against randomness. Finally, Section~\ref{conclusion} concludes the survey.

\section{Background and Preliminaries}
\label{background}
\subsection{AoI fundamentals}
The AoI is a metric introduced to measure the timeliness of information. It is defined as the time elapsed since the generation of the latest update. AoI takes into account both the source's sampling rate and the packet delivery latency, reflecting how fresh the information is at the receiver's end. For example, in a simple queuing system, a low sampling rate results in low packet latency due to the often-empty queue. However, infrequent updates can lead to outdated information at the receiver, thereby increasing AoI. Conversely, increasing the sampling rate does not necessarily improve AoI, as a high arrival rate of updates can cause congestion and significant queuing delays.

\begin{figure}[http]
\centering
\includegraphics[width=0.3\linewidth]{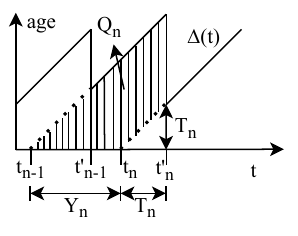}
\caption{Sawtooth-shape age evolution}
\label{AverageAoI}
\end{figure}

% \begin{figure}
% \centering
% \includegraphics[width=0.3\linewidth]{figure/averageAoI.pdf}
% \caption{Sawtooth age waveform.}
% \label{AverageAoI}
% \end{figure}

As shown in Fig.~\ref{AverageAoI}, when a source creates update $i$ with timestamp $u_{i}(t)$, it is sent through the system and eventually arrives at the monitor. The monitor observes its most recent update received at time $t$, with an age of $t - u_{i}(t)$. As time progresses without new updates, the age increases linearly. The average AoI is the time-average of the instantaneous age waveform $\Delta(t)$ over a sufficiently long duration $\tau$, calculated as $\left \langle \Delta \right \rangle _{\tau} = \frac{1}{\tau} \int_{0}^{\tau} \Delta(t)dt$. The interarrival time and system time for the $n$-th update are represented by $Y_{n} = t_{n} - t_{n-1}$ and $T_{n} = t'_{n} - t_{n}$, respectively. Referring to Fig.~\ref{AverageAoI}, the sum of each shaded area, $Q_{n} = \frac{1}{2}(T_{n} + Y_{n})^{2} - \frac{1}{2}T_{n}^{2}$, corresponds to the integral $\left \langle \Delta \right \rangle$. The average AoI is then given by:
\begin{equation}
\Delta = \lim_{\tau \rightarrow \infty} \left \langle \Delta \right \rangle _{\tau} = \frac{E[Q_{n}]}{E[Y_{n}]}. \label{eqn_AoI}
\end{equation}
For detailed analytical explanations of how to apply equation \eqref{eqn_AoI} to various queuing systems, please refer to the survey paper \cite{9380899}.

\subsection{Supervised Learning for AoI Optimization}
Supervised learning is a branch of ML that uses labeled data to train a model that automatically learns patterns and makes predictions. To capture complex representations in data through supervised learning, deep learning (DL) models with many hidden layers are used to learn them via non-linear transformations \cite{kelleher2019deep}.

The role of supervised DL in wireless networking research can be traced to the advancement of ML techniques in supporting complex tasks such as spectrum sensing \cite{zheng2020spectrum, xie2020deep,lee2019deep,umebayashi2017efficient}, link scheduling \cite{cui2019spatial, zhang2019experience, cheng2021deep}, routing \cite{eyobu2023deep, yao2019machine}, modulation/signal classification \cite{o2016convolutional, ye2017power, aslam2012automatic, yu20145, pianegiani2007energy}, interference estimation \cite{chang2015accuracy, assra2015approach,shen2017ica,li2017digital}, and signal/anomaly detection \cite{klassen2012anomaly,parwez2017big, el2009impact,choi2012estimation,li2015joint, zhang2017novel,nguyen2013binary}. A few studies on AoI have leveraged supervised DL in optimizing information freshness \cite{shisher2021age, shisher2022does,leng2021learning, ndiaye2023ensemble,akbari2023constrained, wu2023towards,liu2021age,wang2022deep}.

In a bid to minimize the AoI of energy-harvesting transmitters, \cite{leng2021learning} considered transmitters' status updates as a scheduling problem, accounting for energy and associated interference constraints. A bidirectional long short-term memory (BiLSTM) neural network that maps the system-state input vector to the user-scheduling output vector was proposed to learn a policy for the scheduling problem, with the objective of minimizing the average AoI of transmitters. Similarly, \cite{ndiaye2023ensemble} addressed the problem of minimizing the expected AoI across devices in a UAV-assisted network by proposing an ensemble deep neural network (DNN)-based approach. The ensemble DNN predicts the selection probabilities of users from whom data is to be collected and sent to the server for analysis, thus helping the UAVs dynamically adjust their trajectories to users to minimize the global AoI \cite{ndiaye2023ensemble}.

To avoid the arbitrary sending rates of nodes in a practical collision-based wireless network (e.g., WiFi), \cite{wang2022deep} developed a DL-facilitated AoI optimization approach that integrates service time analysis in 802.11 MAC with AoI queueing analysis to predict the channel condition. The DL model can assist a tagged node in estimating an accurate expected MAC service time, enabling it to adaptively adjust its sending rate to minimize AoI without colliding with other background nodes in the WiFi network. In a device-to-device (D2D) communication setting where interference management is critical, \cite{liu2021age} proposed a DL approach to facilitate link scheduling towards information freshness optimization. The neural network is developed to learn the mapping from device locations to minimum AoI-based link scheduling. Unlike traditional methods, the DL approach in \cite{liu2021age} does not require explicit channel-state information but instead uses a geographic-location-based method (originally proposed in \cite{cui2019spatial}) for AoI-based scheduling in wireless D2D communications.
% \textcolor{blue}{Besides the benefits of DL for AoI minimization in wireless networks, AoI also benefits DL models as described in \cite{wu2023towards} where an AoI-aware federated learning (FL) approach is proposed. In contrast to traditional FL where numerous local clients train their local models with datasets that are unchanging over time, in AoI-aware FL, selected local clients periodically use as fresh as possible datasets to train their local models. In such setting the concept of AoI is leveraged to quantify the freshness of data from selected clients to minimize the global loss and improve the learning accuracy.}

Although supervised DL approaches have demonstrated promising results in AoI optimization, they typically rely on labeled training datasets and may struggle to adapt to dynamic wireless environments with changing channel conditions, traffic patterns, and network states. In contrast, RL does not require pre-collected labeled datasets and can learn adaptive decision-making policies directly through interactions with the environment, using reward feedback to optimize long-term information freshness. These advantages have motivated numerous studies on RL-based AoI optimization in wireless networks, which are systematically reviewed in this survey.

\subsection{Core Concepts of Reinforcement Learning}

RL studies how an agent learns to make sequential decisions by interacting with an environment. In the context of AoI optimization, RL provides a flexible framework for designing adaptive policies that respond to dynamic network conditions and partial system information \cite{712192,zhao2025RLBook}.

\subsubsection{Agent--Environment Interaction and Markov Decision Processes}

An RL problem is commonly modeled as a Markov decision process (MDP), defined by the tuple
$(\mathcal{S}, \mathcal{A}, R, p, \gamma)$.
At each time step $t$, the agent observes a state $s \in \mathcal{S}$, selects an action
$a \in \mathcal{A}$ according to a policy $\pi$, and receives an immediate reward $r \in \mathcal{R}$ after
the environment transitions to a new state $s'$. The transition dynamics are described
by $p(s', r \mid s, a)$, and $\gamma \in [0,1)$ is the discount factor.

The objective of the agent is to maximize the expected return
\begin{equation}
G_t = \sum_{k=0}^{\infty} \gamma^k R_{t+k+1},
\end{equation}
which captures long-term performance.

In freshness-aware systems, the state typically includes freshness metrics, queue status, and channel conditions, while actions correspond to sampling, scheduling, access control, routing, or power allocation decisions.

\subsubsection{Policy and Value Functions}

\begin{itemize}
    \item \textbf{Policy:} A policy $\pi(a|s)$ specifies how actions are selected in each state.

    \item \textbf{State-Value Function:} The state-value function measures the expected
    cumulative discounted reward starting from state $s$ and following policy $\pi$,
    \begin{equation}
        v_\pi(s) =
        \mathbb{E}_\pi \!\left[
            \sum_{k=0}^{\infty} \gamma^k R_{t+k+1}
            \;\Big|\; S_t = s
        \right].
    \end{equation}

    \item \textbf{Action-Value Function:} The action-value function evaluates the expected
    return when taking action $a$ in state $s$ and thereafter following policy $\pi$,
    \begin{equation}
        q_\pi(s, a) =
        \mathbb{E}_\pi \!\left[
            \sum_{k=0}^{\infty} \gamma^k R_{t+k+1}
            \;\Big|\; S_t = s,\, A_t = a
        \right].
    \end{equation}
\end{itemize}

\subsubsection{Bellman Equations and Optimality}
Value functions satisfy recursive Bellman equations, which decompose long-term returns into immediate rewards and expected future values. These equations characterize optimal policies by linking value functions across states and actions, and they form the
theoretical foundation of most RL algorithms \cite{zhao2025RLBook}.
In practice, explicit solutions to the Bellman optimality equations are rarely tractable in large-scale or stochastic systems.

\subsubsection{Dynamic Programming}

Dynamic programming methods, such as value iteration and policy iteration, provide classical approaches for computing optimal policies when the system dynamics are fully known. These methods iteratively refine value functions and policies based on Bellman
updates \cite{712192}. However, their reliance on accurate transition and reward models limits their applicability in complex wireless networks, motivating the use of model-free reinforcement learning methods.

\subsubsection{From Model-Based to Model-Free Methods}

In practical wireless networks, accurate transition and reward models are often unavailable. Model-free RL addresses this limitation by learning directly from observed interactions with
the environment.

\paragraph{Tabular Solution Methods}

Early model-free approaches represent value functions explicitly in tabular form.
\begin{itemize}
    \item \textbf{Monte Carlo methods} estimate returns from complete episodes,
    \begin{equation}
    V(S_t) \leftarrow V(S_t) + \alpha \big[G_t - V(S_t)\big].
    \end{equation}

    \item \textbf{Temporal-Difference methods} update values incrementally using bootstrapping,
    \begin{equation}
    V(S_t) \leftarrow V(S_t) + \alpha \big[R_{t+1} + \gamma V(S_{t+1}) - V(S_t)\big].
    \end{equation}

\item \textbf{On-policy and Off-policy control:} Once value estimates are available, policies can be improved.
On-policy methods, such as SARSA, update the action-value function by following the same policy that is being learned:
\begin{align}
Q(S_t,A_t) &\leftarrow Q(S_t,A_t) + \alpha \Big[ R_{t+1} + \gamma Q(S_{t+1},A_{t+1}) - Q(S_t,A_t) \Big]. \label{eq:sarsa}
\end{align}

Off-policy methods learn the value of the optimal policy while behaving according to a different policy. The most famous example is
Q-learning, whose update rule is
\begin{align}
Q(S_t,A_t) &\leftarrow Q(S_t,A_t) + \alpha \Big[ R_{t+1} + \gamma \max_a Q(S_{t+1},a) - Q(S_t,A_t) \Big]. \label{eq:qlearning}
\end{align}

\end{itemize}

\paragraph{Policy Gradient Methods}

Policy gradient methods directly optimize a parameterized policy. According to the policy
gradient theorem,
\begin{equation}
\nabla J(\theta) \propto
\sum_s \mu_\pi(s) \sum_a q_\pi(s,a)\nabla_\theta \pi(a|s,\theta).
\end{equation}
Actor--critic methods combine policy optimization with value-function estimation, enabling
efficient learning in continuous or high-dimensional action spaces.

\subsubsection{From RL to DRL}

Tabular RL methods do not scale to large or continuous state-action spaces due to memory and
exploration limitations. Deep reinforcement learning (DRL) addresses these challenges by
using neural networks to approximate value functions, policies, or both.

By learning compact representations and generalizing from limited data, DRL enables
RL-based control in complex wireless systems. Representative examples include Deep
Q-Networks (DQN) \cite{Mnih2013PlayingAW} and Proximal Policy Optimization (PPO)
\cite{Schulman2017ProximalPO}.

\subsection{Core concepts of Multi-Agent Reinforcement Learning}
\label{subsec:MARL_POMDP_background}

Wireless networks often involve multiple autonomous nodes, such as sensors, vehicles, UAVs, and terminals, all of which make decisions while interacting within the same physical environment. Multi-agent reinforcement learning (MARL) provides a natural framework for such settings because it extends single-agent RL by allowing each node to act as an independent learner with its own observations and actions. The environment evolves according to the collective behavior of all agents, which creates dynamics that differ significantly from classical single-agent systems \cite{Zhang2019MultiAgentRL,9043893,Gronauer2021MultiagentDR}.

MARL problems can appear in different forms. Some environments encourage full cooperation among agents, others introduce competition, and many involve a mixture of both \cite{marl-book}. Likewise, learning can be organized in several ways. A controller may use global information to train a joint policy, each agent may learn independently using only local information, or training may use global information while execution remains decentralized. These ideas provide a high-level picture of how MARL systems are structured, and their specific use in information freshness optimization will be examined later in Section~\ref{MARL}.

A common difficulty in multi-agent systems is that no individual agent has access to the complete state of the environment. In wireless networks, for example, a node typically observes only its own queue, its local channel conditions, or the sensing outcome in its immediate area. It rarely knows the queues, actions, or intentions of other nodes. This limited visibility means that the underlying decision problem is naturally modeled as a partially observable Markov decision process (POMDP) \cite{marl-book,9372298}.

In a POMDP, the true state of the environment is represented by \( s \in S \), but an agent receives only a local observation. To act effectively, the agent maintains a belief distribution over possible states. The initial belief is written as
\begin{equation}
    b_0^i = \mu ,
\end{equation}
where \( \mu \) is the prior distribution. After the agent takes action \( a_t^i \) and receives observation \( o_{t+1}^i \), the belief is updated according to \cite{marl-book}
\begin{equation}
    b_{t+1}^i(s') \propto 
    \sum_{s \in S} b_t^i(s)\,
    T(s' \mid s, a_t^i)\,
    O_i(o_{t+1}^i \mid a_t^i, s') .
\end{equation}

The update reflects how likely the agent is to transition from state \( s \) to \( s' \) under action \( a_t^i \), and how likely it is to receive observation \( o_{t+1}^i \) in that new state. After normalization, the belief forms a compact summary of the agent's history and provides the information necessary for decision-making. While conceptually straightforward, this filtering process becomes computationally demanding when the environment is large or when many agents interact.

The concepts introduced in this subsection form the essential foundations for understanding the MARL techniques used later in Section~\ref{MARL} to optimize information freshness in decentralized wireless networks.

\subsection{Advantage of RL over Traditional Optimization}

Building on the earlier discussion of RL techniques, its strengths in adaptability, scalability, and efficient policy search naturally align with the unique challenges of AoI optimization in wireless networks. The AoI problem is often formulated as a mixed integer nonlinear programming (MINLP) problem due to the complex, dynamic nature of wireless networks \cite{10433247,9195789,9930881,10130737}. Such problems inherently involve a mix of integer and continuous variables and nonlinear constraints, making them computationally challenging to solve. Traditional optimization approaches for MINLP include exact algorithms, approximate algorithms, and heuristics, each with its own limitations \cite{ZHANG2023205},\cite{DBLP:journals/corr/abs-2003-03600}.

Exact algorithms, such as branch-and-bound \cite{MORRISON201679}, guarantee global optimality but are computationally prohibitive, especially for large-scale problems. Approximate algorithms, while faster, may fail for inapproximable NP-hard problems, which are common in MINLP. Heuristics, on the other hand, offer practical solutions that balance computational efficiency and solution quality, but they lack theoretical guarantees and often require manual design, relying heavily on domain expertise. These methods struggle to adapt to the dynamic and stochastic nature of wireless networks, where system parameters like traffic loads, channel conditions, and resource availability frequently change \cite{ZHANG2023205,Bengio2018MachineLF,DBLP:journals/corr/abs-2003-03600}.

RL overcomes these limitations by introducing a data-driven, heuristic approach that eliminates the need for manual design and can generalize to new problem instances without requiring significant reconfiguration. RL automates the policy search process, adapting dynamically to evolving parameters and leveraging exploration strategies to escape local minima, a limitation of traditional heuristics \cite{Pathak2019SelfSupervisedEV}. Furthermore, RL leverages the Bellman optimality principle, which guides decision-making to maximize long-term cumulative rewards, addressing the critical need for long-term optimization in AoI. The long-term optimization capability of RL can be illustrated using a chess game analogy. In chess, a move that appears optimal in the short term, such as capturing an opponent’s piece, may lead to a strategic disadvantage later, potentially setting up the player for defeat. Traditional optimization methods often focus on such immediate gains under current system parameters, failing to account for their impact on future outcomes. RL, by contrast, evaluates the cumulative effect of decisions, enabling it to avoid such traps. This feature is especially valuable in AoI optimization, where RL dynamically adjusts scheduling, routing, and transmission decisions to maintain data freshness in a long-term manner.

By combining adaptability and scalability, RL excels in handling the high-dimensional state and action spaces in AoI optimization. These spaces are characterized by the number of network nodes, their respective AoI values, channel states, and traffic patterns, as well as decisions on scheduling, routing, and resource allocation. RL, particularly with DRL, employs neural networks to approximate value functions and policies, enabling efficient computation across large-scale, heterogeneous networks, as illustrated in the previous subsection. This makes RL a natural fit for AoI problems, providing robust performance in environments with unknown dynamics, noisy observations, and unpredictable behaviors.

In practice, RL’s ability to iteratively refine its policies based on observed rewards enables it to dynamically adjust scheduling, routing, and transmission decisions through trial-and-error to maintain data freshness under highly dynamic conditions. This adaptability not only improves performance but also ensures long-term robustness, making RL uniquely suited for AoI minimization in wireless networks.

\section{From AoI to Freshness-Aware Optimization}
\label{freshness metrics}

% ============================
% Table 1. Native AoI Metrics
% ============================
\begin{table}
\centering
\caption{Native AoI Metrics}
\label{tab:native}
\renewcommand{\arraystretch}{1.75}

\begin{tabular}{m{2.3cm} m{6.3cm} m{4cm} >{\centering\arraybackslash}m{2.2cm}}
\hline
\textbf{Variant} & \textbf{Definition / Key Idea} & \textbf{Use Cases} & \textbf{Selected Work} \\
\hline
Classical AoI  & Time elapsed since generation of the latest received update. & Baseline freshness measure. & \cite{6195689} \\
Peak AoI (PAoI) & AoI just before each successful reception. & Tracks worst-case information freshness. & \cite{7415972} \\
AoI Distribution 
& Computes the stationary distribution of classical AoI, including moments, distributions, and extreme-value behavior via renewal and queueing analysis. 
& Characterizing full stochastic behavior of native AoI processes. 
& \cite{DBLP:journals/corr/abs-1806-03487,Inoue_2019} 
 \\
\hline
\end{tabular}
\end{table}

% ===============================
% Table 2. Function-based AoI Metrics
% ===============================
\begin{table}
\centering
\caption{Function-based AoI Metrics}
\label{tab:function}
\renewcommand{\arraystretch}{1.75}

\begin{tabular}{m{3.4cm} m{5.6cm} m{3.6cm} >{\centering\arraybackslash}m{2.2cm}}
\hline
\textbf{Variant} & \textbf{Definition / Key Idea} & \textbf{Use Cases} & \textbf{Selected Work} \\
\hline
Age-penalty / Utility Function & Generalized penalty or utility function of AoI. & Captures heterogeneous urgency. & \cite{8000687} \\
Statistical / Risk-sensitive AoI & Probabilistic or CVaR-based penalty. & Risk-aware freshness control. & \cite{10817529} \\
CoUD / VoIU & Cumulative cost or value gained per update. & Quantifies information value. & \cite{8006543} \\
Effective AoI & Function of estimation error instead of time. & Control-oriented AoI. & \cite{8406891} \\
\hline
\end{tabular}
\end{table}

% ============================
% Table 3. Semantic AoI Metrics
% ============================
\begin{table}
\centering
\caption{Application-oriented AoI Metrics}
\label{tab:application}
\renewcommand{\arraystretch}{1.75}

\begin{tabular}{p{3.1cm} p{6.2cm} p{3.3cm} >{\centering\arraybackslash}p{2.2cm}}
\hline
\textbf{Variant} & \textbf{Definition / Key Idea} & \textbf{Use Cases} & \textbf{Selected Work} \\
\hline
On-demand AoI & AoI evaluated only when the destination requests a status update (event-driven freshness). & Energy-limited or request-driven systems. & \cite{9546792} \\
Effective AoI (EAoI) & 
freshness matters only when information is actually used for decision-making. & Request-driven systems, proactive update scheduling, decision-centric freshness. & \cite{8737508}   \\
Relative AoI (rAoI) & how far the receiver lags behind the freshest update at the source. & Measures communication delay impact. & \cite{8988983} \\
Age of Synchronization (AoS) & Time since receiver desynchronized from source. & Cache / database freshness. & \cite{8437927} \\
Age of Version (AoV) & Version gap between source and cache. & Dynamic content caching. & \cite{9771060} \\
Query AoI (QAoI) & AoI sampled at query instants. & Pull-based systems. & \cite{10376462} \\
Age upon Decisions (AuD) & AoI at decision epochs. & Control or decision systems. & \cite{8887253} \\
Age of Task (AoT) & Time until a computation result is returned. & Edge computing / MEC. & \cite{8902529} \\
Peak Age of Task Information (PAoTI) & Age grows until a correct task execution occurs (e.g., correct classification); PAoTI measures the peak age before each successful task outcome. & Task-oriented communications. & \cite{10226176} \\
Age of Incorrect Information (AoII) & AoI counted only when receiver’s info is wrong. & Correctness-aware freshness metric & \cite{9137714} \\
Age of Correlated Information (AoCI) &
AoI that updates only when all task-relevant correlated packets are received. &
Multi-view perception, wireless camera networks. &
\cite{8406914} \\

\hline
\end{tabular}
\end{table}

Building on the fundamentals of AoI introduced in the previous subsection, we now turn to how the notion of freshness has been extended in the literature. The classical formulation of AoI, as defined in \eqref{eqn_AoI}, provides a convenient measure of information timeliness by averaging the age process over time. While this baseline has proven valuable for analytical studies, it captures only one aspect of freshness. In practical systems, relying solely on the average AoI can be misleading, since many applications are more sensitive to occasional deadline violations, operate under strict resource constraints, or depend on how updates contribute to task performance rather than their raw recency.

To address these limitations, researchers have introduced a variety of freshness-aware metrics that extend the basic AoI metric. These metrics, as summarized in Tables~\ref{tab:native}, \ref{tab:function}, and \ref{tab:application}, can be grouped into three broad categories: Native AoI, Application-oriented AoI, and Content-aware freshness. The first retains the raw AoI process and optimizes its statistics, such as the time-average or PAoI. The second transforms AoI through penalty functions, thresholds, or probabilistic constraints to reflect risk sensitivity, deadlines, and resource budgets. The third category incorporates task-level objectives by directly measuring how update staleness affects downstream estimation, prediction, or control performance, often accounting for source correlations or semantic values. In the following, we discuss each category in detail and highlight representative formulations and studies.

\subsection{Native AoI}

The first category includes the original definitions of AoI and PAoI (PAoI). These metrics aim to minimize statistics of the raw age process without further transformation. A representative objective is the time-average AoI, as defined in equation \eqref{eqn_AoI}.
Another common measure is the Peak AoI (PAoI), defined as the maximum age observed just before a packet is received. These formulations were first introduced in \cite{6195689} and later formalized in surveys such as \cite{9380899}.

Building upon these foundational formulations, Kadota et al.\cite{8514816} provided a rigorous discrete-time characterization of AoI and established the expected weighted-sum AoI as a standard optimization objective for wireless broadcast networks. Their work also derived several low-complexity scheduling policies, such as Greedy, Max-Weight, and the Whittle index, which collectively showed that minimizing the average AoI captures a trade-off between packet delay and delivery regularity. More recently, Zhao and Kadota \cite{11044597} extended this framework to settings where the base station has imperfect or delayed AoI knowledge, demonstrating that the classical expected weighted-sum AoI metric remains an effective measure of freshness even under practical wireless constraints. Together, these studies illustrate the maturity and robustness of the native AoI framework as a baseline for freshness optimization. Nonetheless, such native formulations inherently assume homogeneous flows and uniform packet importance.

Native AoI metrics are well-suited to scenarios in which all information updates are equally important and recency alone serves as the primary indicator of freshness. However, they are limited in that they treat all packets equally, ignoring deadlines, resource costs, or application-level priorities.

\subsection{Function-based AoI}

The second category transforms the raw age process through a penalty function $f(\cdot)$, optionally combined with per-flow weights $w_i$. Instead of treating freshness as linear in time, these metrics emphasize specific regions of the age process, such as large staleness values, deadline violations, or tail behavior. A generic form of such an age-penalized metric can be written as
\begin{equation}
\Phi(t) = f\!\big(\Delta(t)\big),
\label{eq:penalized-aoi-ct}
\end{equation}
%\begin{equation}
%\lim_{T \to \infty}\ \frac{1}{T}\,
%\mathbb{E}\!\left[ \int_{0}^{T} f\big(\Delta(t)\big)\, dt \right].
%\label{eq:penalized-aoi-ct}
%\end{equation}
where the choice of $f(\cdot)$ determines which aspects of staleness are emphasized. Quadratic or exponential penalties \cite{8736523} such as $f(x) = x^2$ or $f(x) = e^{\beta x}$ increase the cost of larger ages disproportionately, while threshold-based penalties such as $f(x) = \max\{0, x-\tau_i\}$ or indicator functions \cite{9743441} activate only when deadlines are violated. Service-level guarantees can also be written as probabilistic constraints of the form
\begin{equation}
\Pr\{\Delta(t) > \tau\} \leq \varepsilon,
\label{eq:violation-prob}
\end{equation}
which ensure that the probability of a source’s age exceeding its deadline $\tau_i$ remains below $\varepsilon_i$.

Equations \eqref{eq:penalized-aoi-ct} and \eqref{eq:violation-prob} describe the general way of extending AoI into a penalty-based objective, where the function $f(\cdot)$ can be tuned to emphasize different aspects of freshness. Early studies made this idea concrete by exploring specific forms of $f(\cdot)$ in different network settings. For example, Yavascan et al. \cite{9377549} analyzed slotted ALOHA systems with an age threshold, showing that nodes transmitting only when their AoI exceeds a threshold can nearly halve the average age compared to plain random access. Jiang \cite{9488712} studied large multi-access networks under non-linear penalties such as power and logarithmic functions, proving that optimal scheduling policies remain threshold-based but depend on the specific shape of $f(\cdot)$. Tang et al. \cite{10278713} proposed an $\alpha$–$\beta$ penalty function in short-packet communications, which flexibly captures linear, exponential, and logarithmic costs, thereby linking physical-layer reliability with freshness performance. Zheng et al. \cite{8736523} investigated energy-harvesting systems and derived closed-form results for exponential penalties and violation probabilities $\Pr\{\Delta>\tau\}$, revealing the sensitivity of staleness cost to the balance between update generation and energy arrival rates. Collectively, these works established the foundation of age-penalized metrics by selecting an appropriate penalty function $f(\cdot)$ such as quadratic, exponential, logarithmic, or threshold-based. One can model deadlines, tail risks, or user dissatisfaction with stale data in a mathematically tractable way. This established the groundwork for later RL methods, which treat these function-based costs as objectives and learn good policies when system models are too complex or uncertain for analysis.

Building on these deterministic penalties, several works have incorporated system-level constraints or modified the definition of freshness itself. Leng et al. \cite{8712546} studied an energy-harvesting cognitive radio system and formulated AoI minimization as a POMDP, where the secondary user must satisfy both energy causality and collision-avoidance constraints. Their analysis revealed threshold-based structures in the optimal sensing and updating policies, depending jointly on AoI state and battery level. Liu et al. \cite{10225923} considered HARQ-aided NOMA networks, casting weighted AoI minimization as an MDP solved by a Double-Dueling DQN. By coupling the age cost with power allocation and retransmission control, they proposed a Retransmit-At-Will scheme and showed that the resulting policies exhibit clear threshold behavior, illustrating the interaction between physical-layer mechanisms and freshness objectives. These examples demonstrate how age-penalized formulations can directly integrate resource limitations, reliability requirements, and user-driven freshness definitions.

A parallel line of work has focused on explicitly capturing tail risks and reliability guarantees by modifying the objective function itself. Risk-sensitive metrics such as Conditional Value-at-Risk (CVaR) and Statistical AoI (inspired by entropic value-at-risk, EVaR) provide flexible ways to emphasize worst-case or violation-aware performance. For instance, minimizing the CVaR of the average age can be expressed as
\begin{equation}
\min_{\pi} \; \mathrm{CVaR}_{\alpha}\!\left( f(\Delta(t))\right),
\label{eq:cvar}
\end{equation}
where $\alpha$ controls the level of risk aversion and the randomness arises from stochastic arrivals, channel states, and policy-induced dynamics. Zhou et al. \cite{9149370} showed how CVaR can be embedded into AoI optimization to mitigate rare but severe staleness events. Xiao et al. \cite{10817529} proposed Statistical AoI, an EVaR-inspired metric that smoothly interpolates between average and worst-case peak age, offering a tunable balance between efficiency and reliability. Huang et al. \cite{10462087} addressed AoI-guaranteed bandits, designing online learning algorithms that satisfy strict per-source AoI requirements even under unknown channel reliabilities. These risk-aware metrics are particularly relevant in systems with service-level agreements (SLAs), URLLC demands, or stringent energy and airtime budgets. Collectively, the age-penalized family spans both constraint-integrated and risk-sensitive formulations, extending the classical AoI framework toward practical operation under resource limits and reliability guarantees.

\subsection{Application-oriented AoI}

While function-based AoI metrics reshape the age process by imposing penalties, thresholds, or risk functions, they still treat freshness as a function of time elapsed since the last update. In contrast, application-oriented freshness metrics incorporate application state or inter-source correlations, explicitly tying freshness to whether information is useful for decision-making. This distinction justifies a separate category: the former focuses on cases where staleness is numerically costly, while the latter focuses on cases where it is meaningful for the application. 

As a result, the penalty depends not only on the age but also on task-related signals $e(t)$ and, when relevant, correlations across sources. Rather than being defined purely as a function of elapsed time, application-oriented freshness metrics evaluate staleness through task-aware cost functions of the form
\begin{equation}
\Psi(t) = 
g\!\big(\{\Delta(t)\},\, e(t)\big),
\label{eq:task-aware-metric}
\end{equation}
where $\{\Delta(t)\}$ denotes the collection of age states associated with multiple sources or views, $e(t)$ may represent prediction error, decision loss, or estimation uncertainty, and $g(\cdot)$ modulates the contribution of age using task feedback and inter-source dependencies.

Several representative metrics follow this principle. Hatami et al. \cite{9900377} introduced the notion of On-Demand AoI \cite{9546792} for cache-enabled IoT systems, where freshness is measured only when updates are requested. They minimized on-demand AoI under per-slot energy and transmission constraints, proposing a relax-then-truncate algorithm that achieves asymptotic optimality in large networks. Maatouk et al. \cite{9137714} introduced the \emph{Age of Incorrect Information} (AoII), which penalizes age only when outdated information leads to wrong decisions, making it especially relevant for control and monitoring tasks. Yin et al. \cite{8737508} proposed \emph{Effective AoI} (EAoI), where age contributes only at request times, and developed request-aware scheduling policies using bandit formulations and predictive control. In edge-assisted autonomous driving, Li et al. \cite{Li2021Adaptive} introduced the \emph{Age-of-Result} (AoR), a mobility-aware freshness metric that measures the distance traveled by a vehicle since its last data offloading before receiving the latest computing result. They formulated the computing scheduling problem as a restless multi-armed bandit and developed a Whittle-index-based scheduling policy assisted by deep reinforcement learning to adapt to time-varying vehicle mobility. For correlated sensing, He et al. \cite{8406914} proposed the concept of \emph{Age of Correlated Information} (AoCI) and studied wireless camera networks where an update is fresh only if all correlated packets arrive together, while Yin et al. \cite{9097584} extended AoCI to IoT systems and used DRL with attention and option frameworks to design correlation-aware scheduling.

These content-aware metrics show how freshness can be directly aligned with task performance, such as penalizing incorrect decisions, accounting for request patterns, or capturing cross-source correlations. Hence, bridging communication-layer timeliness with application-level value.

\subsection{Implications}

The three categories of freshness metrics show how the study of timeliness has developed. Native AoI provides a clear, basic way to measure staleness, but it does not capture deadlines, risks, or the actual value of updates. Function-based metrics extend this by adding penalty functions, thresholds, or risk measures to reflect reliability and resource limits. Content-aware metrics go one step further by linking freshness to how information is used in practice, such as AoII, which captures whether outdated data leads to erroneous decisions; EAoI, which accounts for whether updates are explicitly requested; and AoCI, which reflects dependencies among multiple information sources.

For RL, these metrics are important because they define the reward signal. With native AoI, the goal is simply to keep information fresh. With function-based metrics, RL agents can also learn how to handle deadlines, risk, and energy limits. With content-aware metrics, agents can directly optimize freshness for the task at hand, such as sensing, prediction, or control.

In short, the categories move from simple timeliness, to risk- and resource-aware operation, and finally to application-driven objective. This step-by-step view shows why data-driven methods like RL are needed: closed-form solutions often break down in complex or uncertain systems. The following sections build on this foundation to show how RL has been applied to design freshness-aware policies in wireless networks.

\begin{figure*}[htbp]      % placement options: here, top, bottom, page
  \centering
  \includegraphics[width=0.8\linewidth]{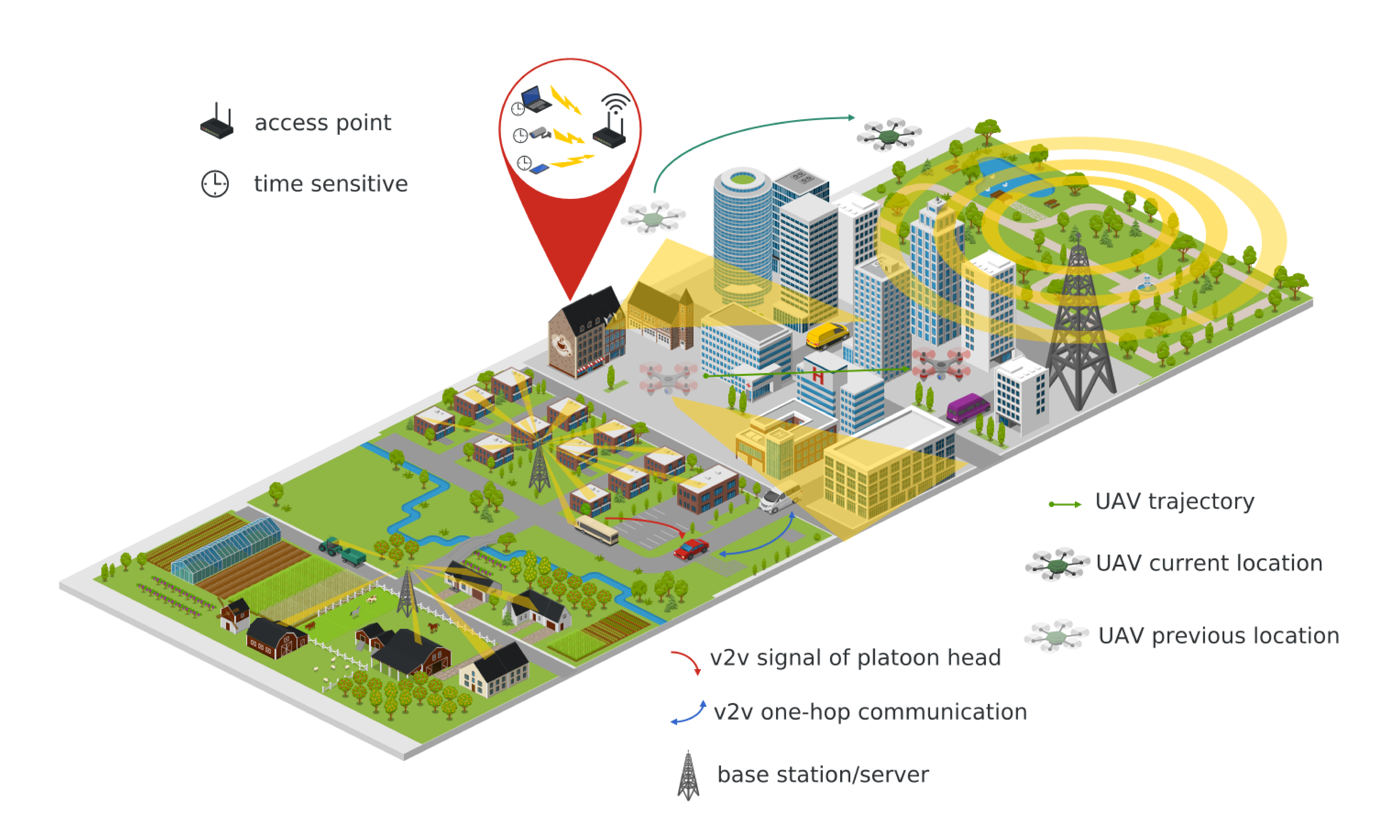}
  \caption{General wireless systems.}
  \label{general system}
\end{figure*}

\section{RL for Freshness: A Policy-Centric Taxonomy}
\label{policy-centric}
The range of freshness metrics discussed in the previous section shows how system requirements have evolved beyond the classical notion of AoI. Each metric highlights a different view of what it means for information to be timely or useful, and these differences naturally lead to different types of decisions that must be learned. This observation suggests that RL methods should be organized according to the kinds of decisions they enable rather than by their algorithmic form.

Guided by this idea, we adopt a policy-centric perspective that reflects the roles RL plays in freshness-aware wireless systems. The first category, \textit{Update-Control RL}, focuses on deciding what information should be generated and when updates should occur, covering problems such as sampling, scheduling, and trajectory selection. The second category, \textit{Medium-Access RL}, concerns how devices compete for and use shared wireless resources in the presence of interference and contention. The third category, \textit{Risk-Sensitive RL}, addresses situations where freshness variability must be controlled by incorporating awareness of rare but costly events or strict operational constraints. These categories align naturally with the different families of freshness metrics introduced in Sec.~\ref{freshness metrics}, and together they provide a coherent framework for understanding how learned policies can support timely information delivery in emerging B5G and 6G wireless systems.

In the following subsections, we describe each policy category in more detail, illustrate its principles, representative models, and its relevance to wireless environments where information freshness is essential.

\subsection{General Wireless Networking System Model}

To set the stage for the RL approaches in the following subsections, we describe a generic wireless networking system model that captures the main challenges of freshness optimization. 

We consider a network with $N$ sources, each generating time-stamped updates for delivery to a common destination (e.g., base station, UAV, or access point). Each source maintains a buffer, and updates compete for transmission over a shared wireless channel with limited resources. The freshness of information from source $i$ at time $t$ is quantified by its AoI, $\Delta_i(t)$, as introduced in Section~\ref{background}. The real application of this general wireless system is shown in Fig.~\ref{general system}.

Fig.~\ref{general system} illustrates several representative communication scenarios that fit within this general model. On the left, a vehicular platoon relies on vehicle-to-vehicle (V2V) links to maintain coordination. The platoon head periodically broadcasts status information, while following vehicles exchange one-hop V2V updates to support cooperative driving and collision avoidance. In the center region, a UAV-assisted network is depicted, in which a UAV collects time-sensitive data and forwards it to an access point or base station. Its trajectory, current position, and previous locations affect link quality and determine how frequently each ground device can transmit fresh updates. On the right, a dense urban setting depicts heterogeneous wireless nodes sharing unlicensed spectrum, such as WiFi or CSMA-based IoT devices. These nodes contend for medium access, and their update performance depends on backoff dynamics, traffic bursts, and congestion around the access point. Although these environments differ in scale and topology, they share a fundamental challenge: multiple sources with time-stamped information must access constrained wireless resources in a timely, coordinated manner. This common structure motivates the policy-centric RL strategies discussed next.

As mentioned earlier, several stochastic factors complicate freshness-aware decision making:
\begin{itemize}
    \item \textbf{Channel uncertainty:} fading, interference, and noise affect whether packets are successfully received.
    \item \textbf{Contention:} simultaneous transmissions may collide or trigger scheduling conflicts under shared access.
    \item \textbf{Energy limits:} many devices operate on batteries or harvested energy, constraining update frequency.
    \item \textbf{Mobility:} in vehicular or UAV-assisted networks, time-varying positions influence link qualities and routing paths.
\end{itemize}

This system can be cast into the RL framework as follows:
\begin{itemize}
    \item \textbf{State:} includes instantaneous AoI values $\{\Delta_i(t)\}_{i=1}^N$, channel conditions, queue occupancies, energy levels, and mobility indicators.
    \item \textbf{Action:} represents node-level control, such as sampling new updates, scheduling transmissions, adjusting access probabilities, or modifying trajectories.
    \item \textbf{Transition:} governed by stochastic packet arrivals, wireless outcomes, contention resolution, and mobility dynamics.
    \item \textbf{Reward:}  reflects information freshness objectives and serves as a
generic abstraction encompassing different freshness metrics discussed in
Section~\ref{freshness metrics}. A commonly used instantaneous formulation is
    \begin{equation}
        r_t = - \sum_{i=1}^N w_i f(\Delta_i(t)),
    \end{equation}
    where $w_i$ are source weights and $f(\cdot)$ is an application-dependent freshness
penalty. For native AoI metrics, $f(\Delta_i)=\Delta_i$, while function-based and
application-oriented freshness metrics are captured through nonlinear, thresholded,
or task-aware choices of $f(\cdot)$. Risk-sensitive formulations further modify how
such per-slot penalties are aggregated over time or across trajectories, rather than
the instantaneous reward itself.
\end{itemize}

The objective is to learn a policy $\pi(a|s)$ that minimizes long-term freshness-related costs under the inherent randomness of the network. This generic model provides a common foundation on which different RL strategies operate: depending on whether the focus is on update generation and scheduling, medium access control, or robustness against tail events, the system's state–action–reward structure is instantiated differently. Building on this foundation, the following subsections examine how RL has been applied across these distinct functional roles.

This unified system model provides the basis for the RL approaches discussed in the remainder of this section. To complement the upcoming subsections, Tables~\ref{tab:rl_aoi_sampling_scheduling}, \ref{tab:rl_aoi_trajectory_scheduling}, and \ref{tab:rl_aoi_protocols} compile representative studies related to the policies examined here. The tables summarize the main elements used in these works, including the state representation, action space, reward design, learning method, and the network setting considered. They are intended to serve as supporting references as we discuss each policy type in detail.

\begin{table}
    \centering
    \caption{Summary of RL for Freshness Optimization through Sampling and Scheduling Policies}
    \label{tab:rl_aoi_sampling_scheduling}
 % --- tighten to fit one page ---
    \scriptsize
    \setlength{\tabcolsep}{3pt}            % default ~6pt
    \renewcommand{\arraystretch}{1.0}     % default 1.0

    \begin{tabularx}{\textwidth}{
        |m{0.6cm}
        |>{\raggedright\arraybackslash}X
        |m{1.6cm}
        |>{\raggedright\arraybackslash}p{3.2cm}
        |m{1.6cm}
        |m{2.2cm}|
    }
        \hline
        \textbf{Paper} & \textbf{State} & \textbf{Action} & \textbf{Reward} & \textbf{RL Algorithm} & \textbf{System Model} \\ \hline
        \cite{8648525} & Current AoI, number of past retransmission attempts & Idle, transmit new packet, retransmit the previous failed packet & AoI and resource usage & SARSA & Single source-destination pair \\ \hline
        \cite{8580701} & Current AoI, number of past retransmission attempts & Idle, transmit new packet, retransmit the previous failed packet & AoI and number of transmission  & Model-based & Single source to multi destinations \\ \hline
        \cite{Zhou2018JointSS} & AoI at the destination, AoI at the queue & Sampling, scheduling & Cost of combined constraints & Model-based & Multiple sources to single destination \\ \hline
        \cite{8845182} & Battery, energy consumption, AoI at both tx and rx, number of retransmissions & Idle, transmit new packet, retransmit the previous failed packet & AoI at the receiver & PG, GR & Single source-destination pair \\ \hline
        \cite{8685524} & The most recent successful transmission event, the timestamp of the event, the timestamp of the packets sent, number of packets in the queue & Scheduling & AoI & DQN, Policy Gradient & Multiple sources to multi-destination \\ \hline
        \cite{8969641} & Age of every sensor, throughput achieved, time spent for downloading & Scheduling & Average AoI and penalties for exceeding AoI threshold & A3C & Multi sources to single destination \\ \hline
        \cite{9085402} & Battery level, AoI, uplink and downlink channel power gain from destination node & WET, Scheduling transmission for certain node & AoI & Proposed DRL with Q-learning & Multiple sources to single destination \\ \hline
        \cite{9215013} & Battery level, AoI, age of the packet at the source node, uplink and downlink transmission power gain & Whether generating the new packet or not, whether transmit, WET & Average AoI & VIA & Single source to single destination \\ \hline
        \cite{9692982} & AoI, transmission time of last packet, throughput achieved & Select certain node to Sample and Transmit & Average AoI and penalties for exceeding AoI threshold & A3C & Multiple sources to single destination \\ \hline
        \cite{Xu2023ReinforcementLF} & AoI, remaining energy, packet size, successful transmission probability & Scheduling & Weighted sum of AoI, energy consumption, cost of replacing battery, penalty for violation & Actor-critic-based algorithm & Multiple sources to single destination \\ \hline
        \cite{inproceedings} & Power allocation, transmission round, AoI & Power allocation, retransmission policy & Expected Weighted Sum of AoI & Double-dueling-DQN & Single source to multiple destinations \\ \hline
        \cite{10622227} & Sender's AoI, receiver's AoI, battery level & Idle, monitor, transmit, monitor and transmit & AoI & Q-learning based & Single source to single destination \\ \hline
        \cite{10663259} & Local channel gains, interference level, remaining load, AoI for links, RIS phase-shift matrix &  Vehicle’s channel assignment, power control, RIS phase-shift matrix & AoI and payload & SAC & Multiple sources to single destination \\ \hline
        \cite{10683037} & Channel conditions, AoI of scheduled devices, handover indicator for uplink, number of links for downlink &  Scheduling & AoI reduction with power constraint & PPO & Multiple sources to single destination and single source to multi destinations \\ \hline
        \cite{10507430} & Remaining computation workload, AoI, time elapsed since last task completion, channel gain &  CPU frequency, transmission power, bandwidth allocation & AoI with bandwidth and energy constraints & Q-learning & Multiple sources to single destination \\ \hline
        \cite{9316802} & Set of candidate vehicles & Selecting the vehicle to receive the forwarding packet to one of its neighboring vehicles & High reward for direct transmission to AP, 0 otherwise & Q-learning & Multiple sources to a single destination
        \\ \hline
        \cite{10333650} & Remaining slots, AoI of each user, remaining number of packets in each queue, channel gain of each link in the current slot & Selecting the specific vehicle to receive the forwarding packet to the candidate set & Reward for no AoI outage, 0 otherwise & SAC & Multiple sources to multiple destinations  \\ \hline
    \end{tabularx}
\end{table}

\subsection{Update-control RL}
\label{task_oriented_rl}
Update-control RL focuses on learning when and what information should be updated in a wireless network. This category primarily covers three classes of policies: \textit{sampling}, \textit{scheduling}, and \textit{trajectory planning}. These policies operate at the application and network layers to ensure timely updates from distributed sources to designated destinations. Optimizing these policies is essential for enhancing information freshness in wireless networks, especially in status-update systems where timely data is critical for control and decision-making. A natural way to formulate such problems is via MDPs, where RL agents learn policies that optimize long-term freshness-related objectives. For example, \cite{8404794} applies a DRL framework to minimize AoI through dynamic scheduling under limited bandwidth.

However, real-world implementations face several challenges. Wireless environments involve transmission errors, fluctuating channel states, and energy constraints. Multi-source networks introduce scheduling conflicts, while mobility and mission-critical tasks impose real-time responsiveness requirements. In the following, we group task-oriented RL approaches into three main classes and discuss representative examples.

\subsubsection{Sampling Policies} \label{sampling}
An essential challenge in AoI minimization is determining when sources should generate fresh status updates, especially under energy and bandwidth constraints. Fixed-periodic sampling can either congest the channel with redundant packets or leave the system with stale data. RL provides a natural solution by adaptively adjusting sampling decisions in response to channel dynamics, energy availability, and network-wide objectives.

One representative study is \cite{8845182}, which considers an energy-harvesting (EH) sensor equipped with hybrid automatic repeat request (HARQ). Here, the cost of both sensing and transmission is explicitly modeled, unlike prior works that ignored sensing energy. The problem is formulated as an average-cost MDP, and two RL methods are explored: GR-learning, a value-based approach, and finite-difference policy gradient, which exploits the threshold structure of the optimal policy. Their results show that policy gradient, guided by structural insights, outperforms GR-learning while remaining robust to unknown energy arrivals and channel statistics. This work highlights the importance of incorporating sensing cost into sampling optimization.

Building on this line, \cite{10547087} focuses on opportunistic sampling for EH sources over fading channels, where the node must decide not only whether to sample but also whether to first probe the channel. They propose a novel two-stage MDP formulation that captures this dual decision process. When the system dynamics (energy arrivals and fading) are unknown, they develop a modified Q-learning algorithm adapted to the two-stage action structure. A key contribution is proving that the optimal sampling strategy follows a threshold structure dependent on both age and probed channel quality. Their results further show that channel probing is beneficial when probing energy is relatively small compared to sampling energy, offering practical insights for low-power IoT deployments.

While the above works focus on single-device optimization, \cite{9691928} extends sampling control to large-scale IoT systems where multiple devices with nonlinear physical dynamics must share limited wireless resources. They formulate a joint problem of optimizing sampling frequency at devices and device selection at the base station to minimize the weighted sum of AoI and energy consumption. To address scalability and decentralization, they propose a distributed RL framework based on QMIX that enables devices to cooperatively approximate global Q-values from local observations. Simulations show significant reductions in both AoI and energy compared to centralized DQN and uniform sampling. This distributed approach is particularly valuable for real-time IoT systems with numerous heterogeneous processes.

Together, these works illustrate the evolution of RL-based sampling: from single EH devices with HARQ and sensing cost, to opportunistic probing-and-sampling under fading channels, and finally to distributed multi-device IoT systems with realistic dynamics. They demonstrate how RL can flexibly adapt sampling strategies across diverse system models while ensuring information freshness under practical constraints.

\subsubsection{Scheduling Policies} \label{scheduling}
Scheduling determines which user or flow should be served at each time slot, subject to resource constraints. Unlike sampling, which focuses on generating updates, scheduling policies determine how to allocate scarce transmission opportunities across multiple sources and destinations. The central challenge lies in balancing fairness in freshness among users while adapting to unreliable channels, correlated information, and multi-hop dynamics. RL has been widely applied to this problem, often introducing specialized techniques to address structural complexity or system-specific requirements.

An early study by Ceran et al.\cite{8648525} considered scheduling in a single-user system with automatic repeat request (ARQ) and hybrid ARQ (HARQ) protocols under an average transmission budget. The problem was cast as a constrained MDP, and an average-cost SARSA algorithm was employed to learn scheduling policies online without knowledge of channel error statistics. A key structural insight from this work is that under ARQ, the optimal policy follows a threshold-type randomized structure, while under HARQ, the scheduler faces a subtle trade-off between retransmitting old packets with higher reliability and sending fresh updates with lower reliability but greater timeliness. The contribution of this study was to show that RL can exploit simple ACK/NACK feedback to learn these threshold-like policies directly in unknown environments.

This line of work was extended to multi-user networks by Ceran et al. \cite{8580701}, who studied a system where a source must decide which user to update over unreliable channels. The scheduling problem was formulated as a restless multi-armed bandit, and RL algorithms were applied to adaptively select users based solely on feedback. The main insight here is that, even in the absence of explicit channel state information, ACK/NACK feedback suffices to learn effective scheduling rules that achieve performance close to the optimal benchmark. This demonstrates the scalability of RL from single-user scheduling to heterogeneous multi-user environments.

A more application-driven perspective was introduced by Yin et al.\cite{9097584}, who studied the concept of Age of Correlated Information (AoCI) for IoT systems where application-level freshness depends on multiple correlated data streams. Standard AoI metrics, which reset upon the arrival of a single packet, cannot capture correlation structures in which decisions rely on joint updates from multiple sources. To address this, they formulated scheduling as an episodic MDP and employed DRL. Their framework integrated two specialized techniques: reward decomposition, which separates correlated contributions to simplify learning, and an attention-integrated relevance network that captures cross-source dependencies. These innovations enabled DRL to produce correlation-aware scheduling policies that significantly outperformed conventional baselines, establishing the importance of modeling correlation in freshness optimization.

Beyond static or centralized systems, vehicular participatory sensing presents a highly dynamic and distributed scheduling challenge. Qin et al. \cite{9316802} investigated this setting, in which vehicles act both as sources and relays to forward updates to roadside infrastructure. To handle mobility and congestion, they proposed a hybrid design that combined Lyapunov optimization for threshold-based sampling with an RL-based forwarding policy. Each vehicle learned relay-selection strategies in a distributed fashion, enabling timely updates while maintaining stability despite rapidly changing topology. The novelty of this approach lay in the integration of learning-based routing with analytical stability guarantees, showing how RL can complement classical control techniques in real-time vehicular networks.

Another direction was explored by Beytur et al.\cite{8685524}, who considered multiple flows served by a single server, a setting known to be analytically intractable. By applying policy gradient and deep Q-learning algorithms, they showed that RL can adapt to stochastic arrivals and delays without requiring explicit traffic models. A particularly striking insight from their study is that RL discovered non-work-conserving scheduling rules in which transmissions are deliberately delayed to achieve lower long-term AoI. This counterintuitive yet effective strategy underscored RL's advantage in uncovering policies beyond the reach of conventional scheduling heuristics.

At the physical layer, Liu et al.\cite{10225923} studied AoI minimization in HARQ-aided non-orthogonal multiple access (NOMA) networks, where scheduling interacts with power allocation and retransmission decisions. They proposed a Double-Dueling DQN architecture to jointly optimize these decisions and introduced a novel “retransmit-at-will” scheme that enables retransmissions to adapt flexibly to instantaneous AoI states. Their results revealed that while conventional user-pairing strategies optimized for throughput, they are not necessarily optimal for AoI, highlighting the need for AoI-specific design principles at the PHY/MAC layers.

Taken together, these works illustrate the broad range of RL-based scheduling approaches, spanning single-user systems with structured threshold policies, multi-user and correlated IoT scheduling, distributed vehicular networks, multi-flow service systems, and advanced physical layer designs. Across these diverse settings, RL methods are often combined with problem-specific techniques such as exploiting threshold structures, decomposing rewards, modeling correlations through attention mechanisms, integrating Lyapunov optimization, and using dueling architectures to better capture the underlying structure of freshness optimization problems. Collectively, these studies demonstrate not only the flexibility of RL for scheduling but also the necessity of tailoring learning frameworks to each system's characteristics rather than relying on generic RL formulations.

\subsubsection{Trajectory Planning}
Trajectory planning with UAVs introduces a unique dimension to AoI optimization, as mobile aerial agents must balance timely data collection with propulsion energy and scheduling constraints. Unlike static sampling or scheduling policies, trajectory planning couples spatial movement with freshness, requiring policies that adapt to dynamic environments, large state spaces, and multi-objective trade-offs. RL, and in particular deep RL, has emerged as a natural solution to capture these spatio-temporal complexities while enabling autonomous decision-making in UAV-assisted networks.

One of the first works to apply deep RL to AoI-aware trajectory design was Abd-Elmagid et al. \cite{9013924}, who studied a UAV-assisted network in which a single UAV collects updates from energy-constrained ground nodes. They formulated the problem as a finite-horizon MDP to minimize the weighted sum of AoI across processes, but the large state space rendered exact dynamic programming infeasible. To overcome this, they proposed a deep RL framework that jointly optimizes the UAV’s trajectory and update scheduling. The key insight of this work was to show that DRL can capture the interaction between battery-constrained ground nodes and UAV mobility, achieving significant reductions in AoI compared with heuristic policies.

Building on this, Zhou et al. \cite{8928091} proposed an online AoI-based trajectory planning algorithm for UAV-assisted IoT networks, where the traffic patterns of ground devices are unknown and time-varying. Their method introduced a pre-training stage using randomized policies to accelerate convergence and avoid poor local optima during online learning. By integrating deep RL with this pre-training strategy, their algorithm could adapt trajectories to heterogeneous traffic dynamics, outperforming traditional delay- or throughput-based UAV path designs. The novelty here is in explicitly modeling the mismatch between UAV motion and the sporadic nature of IoT traffic, and demonstrating that RL can adaptively bridge this gap.

Focusing on energy-aware design, Sun et al. \cite{9426899} addressed the joint optimization of AoI, propulsion energy, and transmission energy in UAV-assisted IoT data collection. They introduced a twin-delayed deep deterministic policy gradient (TD3) framework for trajectory planning, termed TD3-AUTP, that handles the continuous and high-dimensional action space arising from joint control of UAV speed, hovering points, and bandwidth allocation. Their simulations demonstrated superior energy–AoI trade-offs compared to baseline DQN and actor–critic methods, highlighting the role of continuous-control DRL in balancing freshness with energy efficiency.

At the network level, Abedin et al. \cite{9285214} considered multiple UAV base stations (UAV-BSs) navigating in 5G-enabled edge computing environments. The goal is to jointly ensure data freshness and energy efficiency under contextual constraints, such as UAV energy levels and AoI thresholds. They proposed a DQN with experience replay, which allowed UAV-BSs to extract useful trajectory policies in real time from large state spaces. The important insight here is that context-aware trajectory learning enabled UAV-BSs to not only reduce AoI but also achieve over 3\% energy savings compared with greedy baselines, illustrating the practicality of DRL for fleet navigation.
\begin{table}
    \centering
    \caption{Summary of RL for Freshness Optimization through Trajectory Planning Policies}
    \label{tab:rl_aoi_trajectory_scheduling}
    \begin{tabularx}{\textwidth}{|m{0.7cm}|m{4.6cm}|m{2cm}|m{3cm}|m{2cm}|X|}
        \hline
        \textbf{Paper} & \textbf{State} & \textbf{Action} & \textbf{Reward} & \textbf{RL Algorithm} & \textbf{Objective Policy} \\ \hline
         \cite{9013924} & Battery level of ground node, AoI of ground node, location of UAV, the difference between the remaining time and the required time to reach its final location & Movement, scheduling action & AoI & Proposed DRL with Q-Learning & Trajectory planning, scheduling \\ \hline
        \cite{8752017} & Sensor visited at the current time & Next sensor to be chosen & Punishment from high AoI & Q-learning & Trajectory planning \\ \hline
        \cite{9162896} & Projection of UAV on ground, AoI of all sensor nodes at UAV, time difference between the remaining time of UAV and the minimum time to reach the final destination, difference between the remaining energy of UAV and energy required for reaching the destination in the remaining time & Movement, scheduling & Additional reward for reaching destination with a non-zero residual energy, punishment for violating the constraints & DQN & Trajectory planning, scheduling \\ \hline
        \cite{9374448} & Energy levels of the nodes, current time step & Scheduling & Normalized weighted sum of AoI & Proposed DRL & Scheduling \\ \hline
        \cite{9285214} & Current location, target position, average energy efficiency, average AoI & Trajectory coordination & Energy efficiency while penalizing actions that violate constraints related to energy or AoI & DQN & Trajectory planning \\ \hline
        \cite{9426899} & Location of the UAV, current AoI of all nodes, coverage condition & Flight distance, flight speed, flight direction & Combination of average AoI, energy consumption of nodes and propulsion energy of UAV & TD3 & Trajectory planning \\ \hline
        \cite{10508811} & Position of UAV, AoI of ground sensor & Selection of ground sensor, velocity, trajectory & AoI & PPO based & Trajectory planning and scheduling \\ \hline
    \end{tabularx}
\end{table}
Beyond trajectory alone, several works explored the joint design of trajectory and scheduling. Ferdowsi et al. \cite{9374448} developed a neural combinatorial RL (NCRL) framework that integrates convex optimization with deep RL for minimizing normalized weighted AoI (NWAoI). The scheduling component introduced combinatorial complexity that standard RL could not handle efficiently, so they proposed an LSTM-based autoencoder to compress the state space into fixed-size vector representations while preserving spatio-temporal dependencies. This approach not only achieved near-optimal performance but also provided analytic lower bounds on achievable AoI, offering design guidelines for setting node importance weights and UAV speeds. The use of neural combinatorial optimization represents a significant step forward in handling large-scale AoI problems.

Finally, a recent line of work has emphasized the energy–freshness trade-off in UAV navigation, where trajectory planning must balance propulsion energy and timeliness of updates. Besides the work of Abedin et al. \cite{9285214}, complementing this, Sarathchandra et al. \cite{11036251} developed a meta-DRL approach that integrates DQNs with model-agnostic meta-learning, enabling UAVs to adapt their trajectories and scheduling policies to minimize both AoI and transmission power under dynamically varying objectives. These works highlight that RL can not only autonomously plan UAV flight paths but also account for heterogeneous operational constraints—including limited battery budgets, energy harvesting, and time-varying traffic patterns, thereby yielding trajectory policies that effectively balance AoI with overall system efficiency.

In summary, RL-based trajectory planning has evolved from single-UAV formulations with finite-horizon MDPs to multi-UAV, energy-aware, and joint trajectory–scheduling frameworks. Across these works, the key methodological advances include the use of continuous-control DRL (e.g., TD3), pre-training strategies for dynamic IoT traffic, context-aware DQNs for large-scale navigation, and neural combinatorial architectures for joint design. Compared to conventional optimization or heuristic methods, these RL approaches not only reduce AoI but also provide structural insights into UAV mobility and energy management, illustrating their potential for scalable deployment in real-world AoI-sensitive applications.

\subsection{Medium-access RL}

While update control RL primarily addresses high-level decisions such as sampling, scheduling, and trajectory planning, an equally critical challenge arises at the medium access control (MAC) layer: determining how nodes share the wireless medium in real time. Unlike higher-layer strategies that operate on relatively longer timescales, MAC-layer decisions are made at the granularity of individual transmissions, where collisions, interference, and energy consumption directly affect information freshness. Traditional protocols such as ALOHA or CSMA/CA rely on static or heuristically tuned access rules, which often fail to adapt to highly dynamic traffic and channel conditions. By framing medium access as a sequential decision problem, RL enables nodes to learn adaptive access strategies that minimize collisions, adjust contention parameters, and balance energy efficiency against timeliness. In this sense, access-oriented RL complements higher-layer strategies by providing intelligent, fine-grained control at the MAC level, ensuring that the shared communication medium is utilized efficiently to support AoI-sensitive applications.

These applications primarily addressed the strategic allocation of resources and paths within relatively structured environments. However, the MAC layer operates at a more granular level, dealing directly with real-time access to the communication medium. This distinction introduces several unique challenges and opportunities for RL: 

1) \textbf{Real-time medium access}: Unlike scheduling and routing, which often operate on a broader timescale, the MAC layer requires immediate decisions about which nodes can access the medium at any given moment. RL algorithms must operate efficiently in real time to adapt to fluctuating conditions and maintain network performance.

2) \textbf{Collision avoidance}: At the MAC layer, preventing collisions is paramount. Traditional approaches like CSMA/CA use fixed strategies, while RL offers a dynamic method to predict and avoid potential collisions based on ongoing learning from network behavior.

3) \textbf{Fine-grained control}: While trajectory planning and routing focus on optimal paths and resource distribution, MAC protocols require fine-grained control over access mechanisms. RL can dynamically adjust parameters such as backoff times and access probabilities to optimize throughput and latency.

4) \textbf{Energy efficiency}: Energy constraints are critical at the MAC level, especially in sensor networks and IoT devices. RL can learn energy-efficient access strategies that prolong the network's lifetime without compromising performance.

\begin{table}
    \centering
    \caption{Summary of RL for Freshness Optimization in MAC}
    \label{tab:rl_aoi_protocols}
    \begin{tabularx}{\textwidth}{|m{0.7cm}|m{3.5cm}|m{2.2cm}|m{3.2cm}|m{2.2cm}|X|}
        \hline
        \textbf{Paper} & \textbf{State} & \textbf{Action} & \textbf{Reward} & \textbf{RL Algorithm} & \textbf{MAC Component} \\ \hline
       \cite{9386228} & AoI at terminal’s queue, AoI at destination & Probability & Utility minus transmission tax for transmission; no tax if silent & DDPG & Access mechanism ($p$-persistent CSMA) \\ \hline
        \cite{10020660} & Current urgency order, transmitting packet age, previous action, feedback, buffer state & Transmit, stay silent & Difference between age of transmitted and updated packets & PPO & Access mechanism (slotted ALOHA) \\ \hline
        \cite{Li2023LearningCW} & Packet collision probability & Congestion window parameter & Average AoI (random sampling) & Double DQN & Access mechanism (CSMA/CA) \\ \hline
        \cite{10652761} & Average peak AoI, AoI of user selected by AP after collision & Mini-slot scheduling & Reduced AoI & SAC & Access mechanism (slotted ALOHA) \\ \hline
        \cite{10499046} & AoI in buffer, AoI at destination, buffer state, AoI threshold indicator & Transmit, stay silent & Reduced AoI with penalty for excessive AoI & DQN based & Access mechanism (slotted ALOHA) \\ \hline
    \end{tabularx}
\end{table}

\begin{figure}[http]
\centering
\includegraphics[width=1.0\linewidth]{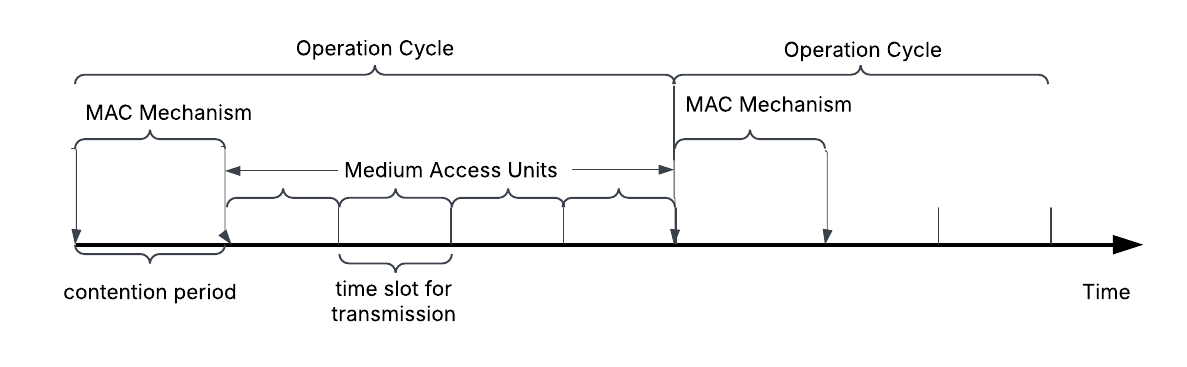}
\caption{The architecture of MAC components from \cite{article1}.}
\label{RL for MAC Protocol}
\end{figure}
A number of works have explored how RL can be integrated into MAC protocol design. As summarized in Fig.~\ref{RL for MAC Protocol}, the MAC reference model in \cite{article1} and \cite{article2} has been extended with RL-based mechanisms across three categories. First, \textit{operation cycle optimization} exploits RL to adapt the timing and structure of MAC cycles to current traffic demands. For example, the eOQ-MAC protocol employs Q-learning to manage superframes for emergency transmissions, striking a balance between packet loss and latency \cite{8986096}. Second, \textit{dynamic medium access units} use RL to adjust frame sizes or slot allocations; RL-IRSA \cite{8920122} is a representative protocol that dynamically selects access units, ensuring efficient channel utilization under varying network loads. Third, \textit{advanced MAC mechanisms} leverage deep RL to refine contention-based access. A notable example is Neuro-DCF \cite{10.1145/3466772.3467043}, which employs DRL to optimize backoff parameters in CSMA-like protocols, thereby reducing collisions and improving throughput.

In the context of AoI optimization, several studies have directly targeted RL-enhanced MAC mechanisms. A DRL-based scheme was proposed in \cite{10020660} to minimize AoI in slotted ALOHA systems designed for massive machine-type communication. The work in \cite{9386228} addressed $p$-persistent CSMA, enabling nodes to learn contention probabilities that jointly minimize collisions and information staleness. Another study \cite{9798137} extended this approach to dynamic environments where nodes may join or leave the network, with RL agents adapting access decisions in real time. More recently, \cite{Li2023LearningCW} considered IEEE 802.11 networks and introduced a DRL-based framework to dynamically learn contention window (CW) sizes under varying traffic loads, directly optimizing AoI while maintaining fairness and throughput. Another key challenge in access-oriented AoI optimization is dealing with unreliable channels. In this regard, \cite{8648525} studied retransmission-based protocols, comparing ARQ and HARQ under uncertain error statistics. They showed that an average-cost SARSA algorithm can gradually learn channel behaviors from ACK/NACK feedback, enabling adaptive retransmission policies that avoid wasting resources on outdated packets. This result highlights the advantage of RL in learning from minimal feedback at the MAC layer, which is often the only information available for real-time adaptation.

Collectively, these works demonstrate that RL provides a versatile toolkit for enhancing MAC protocols. By learning operation cycles, adapting contention parameters, and intelligently handling retransmissions, RL-based designs achieve lower AoI and higher efficiency compared to traditional fixed-access methods. Access-oriented RL thus bridges the gap between high-level scheduling policies and the physical realities of wireless access, ensuring that freshness is preserved even under dense traffic and unreliable channels.

\subsection{Risk-Sensitive RL}
While update control and medium access RL methods have shown strong potential for improving information freshness through intelligent sampling, scheduling, trajectory planning, and channel access, most existing approaches formulate their objectives in terms of expectations, such as minimizing long-term average freshness metrics. Although effective at capturing mean system behavior, expectation-based formulations do not reflect the tail behavior of freshness dynamics. In many real-world systems, including mission-critical IoT, ultra-reliable low-latency communications, and safety-critical vehicular or industrial networks, rare but extreme degradations in freshness can be far more damaging than small fluctuations around the average. For example, while modest reductions in average freshness may be acceptable in non-critical monitoring applications, violating freshness thresholds in remote healthcare, haptic control, or autonomous driving can lead to serious system failures.

This limitation motivates the development of risk-sensitive RL, which departs from conventional freshness optimization by redefining the learning objective itself. Rather than optimizing only the expected freshness, risk-sensitive RL explicitly accounts for reliability, threshold violations, and worst-case performance. In contrast to update control RL, which focuses on learning what and when to update, or medium access RL, which determines how nodes share wireless resources, risk-sensitive RL incorporates risk-aware criteria such as violation probabilities, statistical freshness metrics, utility-based penalties, or constrained formulations like CMDPs directly into the objective. This shift enables learning policies that provide stronger guarantees on tail performance and overall system reliability.

Recent studies have shown that RL can be adapted to these risk-sensitive formulations through specialized techniques, including threshold-penalized rewards, safe DRL with violation guarantees, distributional RL for capturing full AoI distributions, and multi-agent architectures that minimize threshold-violation ratios in decentralized settings. By embedding these risk-sensitive objectives, RL agents can move beyond average-case optimization to deliver robust performance guarantees under diverse constraints, such as energy causality, unreliable channels, and strict latency requirements.

In what follows, we review representative works in this emerging area, highlighting how risk-sensitive RL provides a crucial step toward bridging the gap between average AoI optimization and the stringent reliability needs of mission-critical and ultra-reliable networks.

A representative work in this direction is \cite{9692982}, which considers mission-critical IoT monitoring where exceeding freshness deadlines can lead to severe system failures. The authors formulate an optimization problem that combines two objectives: minimizing the average AoI across sensors and minimizing the probability that AoI exceeds a predefined threshold. They develop a deep RL solution based on the actor–critic framework, demonstrating that the proposed method can adaptively balance timeliness and reliability under fluctuating traffic and channel conditions. By directly incorporating threshold violation penalties into the learning process, this study illustrates how risk-sensitive objectives can be naturally embedded in RL formulations for AoI.

In a similar spirit, \cite{8969641} investigates ultra-reliable low-latency communication (URLLC) systems, where extreme reliability constraints require attention to worst-case AoI behavior. The authors introduce an RL scheduling algorithm that minimizes a weighted sum of expected AoI and the probability of AoI threshold violations across multiple sensors. Using an actor–critic training approach over real bandwidth traces, they show that the learned policy reduces both average age and violation probability compared to baseline scheduling rules. This work highlights the importance of explicitly accounting for the tail distribution of AoI in RL-based formulations, making it highly relevant to URLLC deployments.

Several more recent works strengthen this line of research by introducing novel risk-aware mechanisms. De Sombre et al. \cite{10283567} addressed the issue of catastrophic staleness by proposing Q+RS, a variant of Q-learning that explicitly penalizes ``risky states'', defined as situations where the AoI exceeds a safety threshold. By integrating this penalty into the reward function, the algorithm achieves a balance between minimizing average AoI, conserving transmission energy, and reducing the frequency of severe violations. A key technical contribution is the introduction of a \textit{threshold family} approach that is independent of the underlying state-space size. This innovation ensures that the method remains computationally tractable even as the AoI state space grows, offering a scalable way to handle risk-awareness in large systems. 

Building on this foundation, Wang et al. \cite{10181012} extended risk-sensitive RL to decentralized multi-agent settings in UAV-assisted sensing networks. Their objective was to explicitly minimize the AoI violation ratio under energy constraints, a task complicated by the fact that each UAV must coordinate with others in dynamic environments. To achieve this, they incorporated a Transformer-based module (GTrXL) that captures temporal dependencies across UAV observations, enabling agents to better anticipate future violations. Moreover, they introduced an intrinsic-reward mechanism that incentivizes exploration and prevents premature convergence to suboptimal cooperative behaviors. Together, these innovations highlight how advanced neural architectures can enhance robustness and coordination in decentralized risk-sensitive learning tasks.

At the physical layer, Sheikhi et al. \cite{SHEIKHI2025850} investigated UAV networks operating over mmWave channels with energy harvesting, where both channel dynamics and energy availability are highly uncertain. They proposed a \textit{two-timescale risk-sensitive RL algorithm} that simultaneously adapts to fast-varying channel conditions and slower energy arrival processes, while directly incorporating AoI dynamics into the learning objective. The insight here lies in coupling risk-awareness with energy causality, ensuring that learned policies remain not only fresh but also feasible given energy limitations. This dual adaptation mechanism makes the approach especially suitable for real-time deployments in volatile environments. 

More recently, Reyhan et al. \cite{reyhan2025safedeepreinforcementlearning} investigated wireless networked control systems in which PAoI thresholds must be guaranteed in the presence of finite-block-length effects and strict reliability requirements. By embedding control-theoretic constraints such as the maximum allowable transmission interval and delay directly into the DRL training process, they designed a safety layer that projects unsafe actions back into the feasible space. This approach effectively bridges the gap between control-theoretic reliability guarantees and RL-driven communication policies, ensuring that agents learn strategies that are both effective and provably safe.

Beyond reliability guarantees, risk-sensitive AoI optimization can also emerge in energy-limited environments, where exceeding energy constraints poses a fundamental risk to system operation. For instance, \cite{9085402} studied RF-powered communication systems and developed an RL framework that accounts for both AoI minimization and energy causality constraints. Similarly, \cite{9546792} examined status update control in networks with energy-harvesting sensors and cache-enabled edge nodes, proposing Q-learning and value-iteration methods to manage the trade-off between on-demand freshness and finite battery budgets. While these studies do not explicitly model AoI threshold violation probabilities, they embody the broader notion of risk-sensitive design by addressing scenarios in which resource depletion constitutes a critical risk alongside timeliness.

In terms of tail-related metric studies, prior work provides important context for risk-sensitive design. Huang et al. \cite{10462087} explored AoI-constrained bandits with hard guarantees, demonstrating that online learners can satisfy per-source AoI requirements with probability one. Xiao et al. \cite{10817529} introduced the notion of \textit{Statistical AoI}, a risk-aware metric inspired by entropic value-at-risk (EVaR), which unifies average, tail, and violation-aware perspectives into a single formulation. This provides a principled way to define risk-sensitive objectives that can be embedded into RL models. He et al. \cite{10.1109/TMC.2024.3370101} further demonstrated that CMDP-based RL can enforce AoI constraints via Lagrangian dual methods, showing that constrained formulations offer an alternative to risk-penalty rewards.

Taken together, these contributions demonstrate a clear evolution: from single-agent schedulers that penalize risky states, to multi-agent UAV systems leveraging Transformer-enhanced coordination, to physical-layer solutions that integrate energy harvesting and mmWave uncertainty, and finally to safe DRL frameworks that enforce explicit reliability guarantees. The addition of theoretical advances such as Statistical AoI and CMDP-based formulations further enriches the toolkit for risk-sensitive optimization. Collectively, these works show that risk-sensitive RL is not simply about reducing average age, but about developing policies that remain robust against tail events, resource outages, and safety-critical constraints—making it indispensable for AoI optimization in ultra-reliable and mission-critical networks.

\begin{table*}[!htbp]
    \centering
    \caption{Summary of MARL for Freshness Optimization}
    \label{tab:rl_frameworks}
    \begin{tabularx}{\textwidth}{|m{0.7cm}|m{2.5cm}|m{3cm}|m{3.5cm}|X|}
        \hline
        \textbf{Paper} & \textbf{RL Algorithm} & \textbf{Type of the Framework} & \textbf{Type of the MARL Agent} & \textbf{Objective Policy} \\ \hline
        \cite{9013454} &  DRQN based & Fully decentralized & Cooperative & Scheduling and power control policy
 \\ \hline
        \cite{9322539} & DDPG & CTDE & Mixed & Trajectory planning \\ \hline
        \cite{9154432} & Actor-critic based & Fully decentralized & Cooperative & Trajectory planning
\\ \hline
        \cite{9691928} & QMIX & CTDE & Cooperative & Sampling and device selection\\ \hline
        \cite{9386228} & DDPG & Fully decentralized & Mixed & Access Mechanism
 \\ \hline
        \cite{10012697} & DQN & Fully decentralized & Competitive & Trajectory planning and scheduling
 \\ \hline
        \cite{9837720} & Actor-critic based & Fully decentralized & Cooperative & Trajectory planning
 \\ \hline
        \cite{9426913} & Actor-critic based & CTDE & Cooperative & Trajectory planning, scheduling and BW allocation
 \\ \hline
        \cite{9860365} & DQN based & Fully decentralized & Cooperative & Trajectory planning and power control\\ \hline
        \cite{10293964} & DQN & ALL & Cooperative & Trajectory planning and scheduling
 \\ \hline
        \cite{9950310} & DQN & Fully Decentralized & Cooperative & Trajectory planning and scheduling\\ \hline
        \cite{10200769} & QMIX based & CTDE & Cooperative & Trajectory planning and scheduling \\ \hline
        \cite{10077432} & MADDPG based & CTDE & Mixed & Scheduling and power control
 \\ \hline
        \cite{10731760} & TD3 based & CTDE & Cooperative & Trajectory planning, power control and scheduling policy
 \\ \hline
        \cite{10570857} & PPO based & Fully decentralized & Cooperative & Trajectory planning and scheduling
 \\ \hline
        \cite{10540629} & PPO & CTDE & Cooperative & Sampling
 \\ \hline

    \end{tabularx}
\end{table*}

\section{Multi-Agent Reinforcement Learning for Freshness }
\label{MARL}
As wireless networks evolve toward large-scale, distributed, and heterogeneous systems, many freshness optimization problems inherently become multi-agent. In such environments, the information freshness of each device is no longer determined solely by its own actions, such as sampling or scheduling. Instead, it emerges from interdependent decisions made by multiple nodes sharing spectrum, relaying information, or jointly sensing an environment. These interactions cause each agent’s information freshness to depend on the collective behavior of all other agents, creating inherently coupled decision processes. This interdependence naturally motivates the use of MARL.

In this section, we revisit MARL through the lens of information freshness, emphasizing why freshness-centric problems pose unique challenges, how MARL models are formulated for freshness optimization, and what algorithmic patterns have emerged in recent literature across wireless access, cooperative sensing, multi-hop routing, and edge computing.

Table~\ref{tab:rl_frameworks} summarizes representative MARL studies applied to information freshness optimization. It lists the learning method, the underlying coordination framework, the type of interaction among the agents, and the corresponding objective policies considered in each work. This table serves as a supporting reference for the architectural and algorithmic discussions that follow.

\subsection{Centralized vs. Decentralized System Architectures}
Information freshness optimization often involves multiple agents whose actions affect one another. Because freshness evolves through the collective behavior of the network, the choice of system architecture directly affects how effectively multi-agent learning can be used. Centralized, decentralized, and hybrid designs each offer different advantages and limitations when applied to freshness-critical wireless systems.

\begin{figure}[htbp]
\centering

\begin{subfigure}[t]{0.49\linewidth}
  \centering\vspace{0pt}
  \includegraphics[page=2,width=\linewidth]{figure/survey_for_MARL_figures.pdf}
  \caption{Centralized architecture. The controller takes observations, actions, and rewards from all agents to train and send commands.}        \label{Centralized}

\end{subfigure}\hfill
\begin{subfigure}[t]{0.49\linewidth}
  \centering\vspace{0pt}
  \includegraphics[page=1,width=\linewidth]{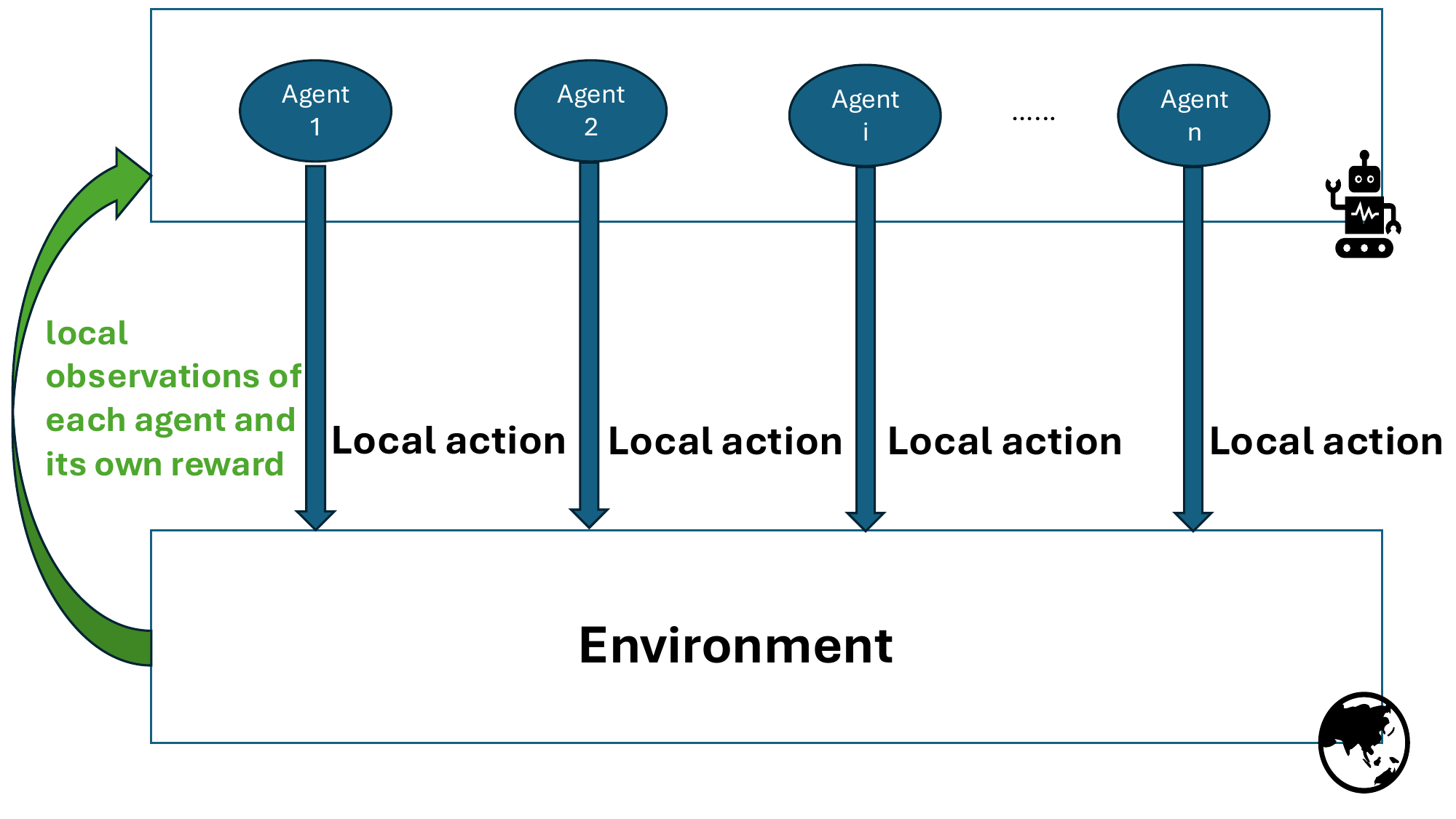}
  \caption{Fully decentralized architecture. Agents act independently using only local observations and rewards.}
  \label{FullyDecentralized}

\end{subfigure}

\medskip

\begin{subfigure}[t]{0.75\linewidth}
  \centering\vspace{0pt}
  \includegraphics[page=3,width=\linewidth]{figure/survey_for_MARL_figures.pdf}
  \caption{CTDE architecture using actor-critic as an example. The centralizer has \( n \) critic networks, and each agent is an actor network. During training, each critic network takes actions and observations from all agents and the reward of its corresponding agent as inputs to compute the Q-value.}        
  \label{CTDE}

\end{subfigure}

\caption{Comparison between centralized, fully decentralized, and CTDE architectures.}
\label{CombinedFig}
\end{figure}

In fully centralized systems, a central controller observes the global network state and makes joint decisions for all nodes, as illustrated in Fig.~\ref{Centralized}. This architecture allows learning algorithms to exploit complete information about AoI states, queue backlogs, channel conditions, and resource availability across the network. For information freshness optimization, centralized control can achieve high performance by coordinating updates, scheduling, and access decisions in a globally consistent manner, reducing redundant transmissions and balancing freshness among competing sources. Centralized learning is particularly effective when the network scale is moderate and communication overhead is acceptable, since it avoids the nonstationarity and partial observability challenges inherent in decentralized multi agent learning. However, the reliance on global state collection and a single decision maker introduces scalability limits, communication overhead, and vulnerability to single point failures, which become critical concerns in large scale and highly dynamic wireless systems.

In contrast, fully decentralized systems empower individual nodes to make decisions based solely on local information, as shown in Fig.~\ref{FullyDecentralized}. This structure naturally improves scalability by eliminating the need for a single controller and distributing decision-making across the entire network. Each node operates independently, which enhances robustness and minimizes the risk that a single failure disrupts the entire system. Decentralized systems are also well-suited for environments where rapid reactions to local channel or traffic changes are essential for maintaining information freshness.

The main challenge in decentralized operation is that each agent observes only part of the environment. This limited visibility makes coordination difficult and can lead to collisions or redundant transmissions, thereby degrading overall freshness. In other words, the decision problem becomes partially observed. In \cite{9013454}, the authors study a wireless ad hoc network with no centralized infrastructure, so all decisions must be made in a distributed manner. To handle partial observability, they propose a multi-agent reinforcement learning method that uses a DRQN to extract useful information from past observations.

A promising middle ground is the centralized training with decentralized execution approach, illustrated in Fig.~\ref{CTDE}. During training, a central controller has access to global information and uses it to guide the learning process. This helps the agents learn coordinated and stable policies, even though the environment is highly coupled through information freshness dynamics. Once training is complete, the agents operate independently and rely only on their own observations. This preserves scalability and removes the need for real-time centralized control.

An example of this idea appears in \cite{9322539}, which studies a system of multiple UAVs that aim to maintain information freshness for ground users. During training, all UAVs send their states to a base station, which then uses this complete view of the network to train each UAV’s policy under a DDPG framework. After training, the UAVs operate autonomously. Each UAV decides its movement and transmission pattern using only local information, such as its own position, its current freshness requirement, and the interference level it experiences. This design combines the advantages of central learning with those of independent execution in the field.

Further comparisons of architectural choices are presented in \cite{10293964}. In the centralized RL framework, all UAVs depend on a base station to compute optimal policies using the full network state. While this produces high-quality decisions, the method becomes too complex for larger networks. In the cooperative multi-agent framework, UAVs train their own policies while exchanging selected pieces of information with one another. This improves overall freshness but requires heavy communication among the agents. A partially cooperative framework reduces this burden by sharing only essential freshness information through a base station. It offers a good balance by lowering communication overhead while still enabling a certain degree of coordination. Finally, in the fully decentralized framework, each UAV acts entirely on its own observations. This yields the simplest execution model and very low communication cost, but it usually produces weaker performance because agents cannot coordinate their actions.

For readers interested in a broader discussion of RL architectures, additional background is available in \cite{Zhang2019MultiAgentRL,9043893,Gronauer2021MultiagentDR}.

\subsection{MARL Scenarios}
The architectures introduced earlier describe how learning can be distributed across agents, but they do not fully capture how agents interact with one another when the goal is to maintain information freshness. In wireless networks, every action that affects the channel, interference level, or sensing coverage can immediately alter the freshness experienced by other agents. As a result, MARL scenarios are best understood by examining the relationships among agents rather than focusing solely on the system architecture. Freshness optimization can therefore be categorized into three broad interaction settings: fully cooperative, fully competitive, and mixed cooperative–competitive.

\textbf{Fully Cooperative Agents:}
In many freshness-aware systems, all agents share the same objective: improving the freshness of information available to a base station, a fusion center, or a set of users. Because one agent’s outdated update can degrade the system-wide freshness, cooperation becomes necessary to achieve stable and efficient operation. Cooperative MARL offers a natural framework for these systems, since agents can learn to coordinate transmission schedules, sensing tasks, or movement patterns to maintain low global freshness.

For example, in \cite{10012697}, multiple UAVs jointly minimize the AoI by learning coordinated trajectories and communication schedules using a multi-agent DQN scheme. A similar cooperative design appears in \cite{9154432}, where UAVs share a common reward that reflects the overall freshness reduction of the system. This shared reward aligns individual actions with the global freshness objective, encouraging UAVs to select tasks and routes that collectively reduce information staleness. Cooperation can also be enhanced through attention mechanisms, as shown in \cite{9837720}, which allow UAVs to focus on the most relevant features of the environment when working together to update information efficiently.

\textbf{Fully Competitive Agents:}
Freshness optimization can also create competition, especially when agents attempt to maintain low freshness for themselves while sharing a limited communication resource. A node that transmits aggressively may succeed in reducing its own information staleness, but at the cost of increasing collisions and degrading freshness for others. This inherently competitive interaction appears in networks with independent entities or selfish nodes, where preserving individual freshness is prioritized over system-wide performance. In these cases, MARL is used to find equilibria in which agents adopt stable strategies that respond to each other’s behavior without centralized coordination.

\textbf{Mixed Cooperative and Competitive Agents:}
In many practical systems, agents exhibit both cooperative and competitive tendencies. They may need to cooperate to avoid excessive collisions, yet they may still act competitively to lower their own freshness. Such mixed settings arise naturally in decentralized wireless networks, including contention-based MAC protocols or multi-domain IoT deployments.

A representative example is found in \cite{9386228}, which studies a $p$-persistent CSMA network where each node independently decides whether to transmit and seeks to minimize its own AoI. The authors introduce a “transmission tax” to discourage over-aggressive transmission behavior. Although nodes do not explicitly coordinate, the tax implicitly promotes cooperative behavior by reducing the likelihood of persistent collisions. This helps maintain fresher information for the system as a whole, demonstrating how mixed cooperative–competitive interactions shape the design of MARL strategies for freshness.

These MARL scenarios illustrate that the structure of agent interactions fundamentally shapes how freshness can be optimized. Whether agents share a common freshness goal, compete for individual freshness, or operate in between, understanding these relationships is key to selecting appropriate MARL algorithms and designing stable policies for decentralized wireless networks.

\subsection{Structural Aspects of Freshness-Aware MARL}

%Although MARL offers a powerful framework for modeling interactions among multiple wireless nodes, real deployments introduce several difficulties that do not appear in traditional single-agent learning. Two challenges are especially important for information freshness: partial observability and non-stationarity. Both can significantly limit an agent's ability to learn reliable policies that maintain low freshness across the network \cite{marl-book}.

Although MARL provides a powerful framework for modeling interactions among multiple wireless nodes, real deployments exhibit structural characteristics that are absent in traditional single-agent learning. In freshness-aware systems, two characteristics are particularly prominent: partial observability and non-stationarity. These characteristics affect how agents infer network conditions and adapt their policies over time, thereby influencing system-wide information freshness \cite{marl-book}.

\textbf{Partial Observability:}
In decentralized wireless networks, agents typically lack a complete view of the environment. A sensor typically knows only its own queue and local channel quality. A UAV performing cooperative sensing observes its own measurements or its relative position but cannot directly infer the actions or intentions of neighboring agents. These limited and noisy observations make it difficult to understand how the actions of one agent influence freshness elsewhere in the network. Because freshness depends on coordinated timing among agents, the lack of visibility can easily lead to suboptimal actions such as redundant updates or unexpected collisions.

\begin{figure}[t]
\centering
\includegraphics[width=0.5\linewidth]{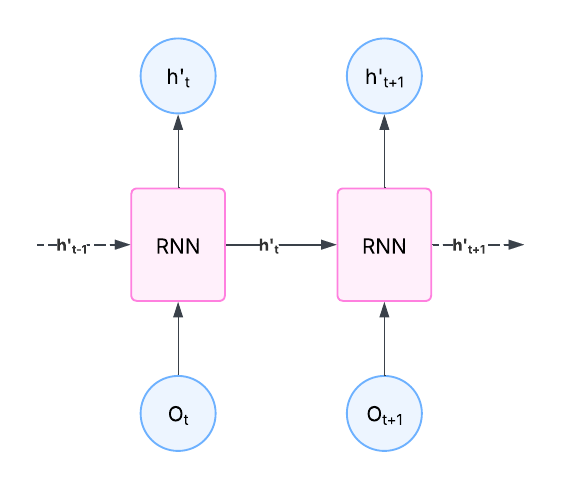}
\caption{RNN for belief update}
\label{RNN}
\end{figure}

To address this limitation, MARL systems often rely on methods that extract information
from sequences of observations rather than depending on explicit knowledge of the global
state. Maintaining exact belief distributions is possible in theory \cite{9372298}, but
it becomes computationally infeasible when many agents interact. In practice, neural
network architectures that incorporate temporal memory are more commonly adopted.
As illustrated in Fig.~\ref{RNN}, recurrent neural networks process streams
of observations over time and maintain hidden states that summarize past information.

Deep Recurrent Q Networks (DRQN) \cite{Hausknecht2015DeepRQ}, as used in \cite{9013454},
and LSTM-based agents, such as the one in \cite{9085402}, learn internal representations
of past observations that serve as implicit beliefs about the environment. These learned
representations enable agents to make decisions under uncertainty without requiring
explicit belief updates or full knowledge of system dynamics. For completeness, point-based
value iteration (PBVI) \cite{10.5555/1630659.1630806} remains a classical approximation
method for belief updates, although its computational overhead limits its use in large
multi-agent networks.

\textbf{Non Stationarity:}
A second challenge arises because all agents are learning simultaneously. As each agent adapts its policy, the environment observed by the others changes as well. A strategy that works at one moment may stop working once nearby nodes adjust their behavior. For information freshness, where timely coordination is essential, even small changes in a neighbor's policy can cause rapid shifts in the overall freshness dynamics. These fluctuations make it difficult for individual agents to predict the consequences of their actions, often slowing or destabilizing the learning process.

Several techniques have been proposed to reduce the impact of non-stationarity. One influential idea is CTDE as illustrated previously. A centralized critic can observe joint information during training, allowing it to account for how agents influence one another. During execution, however, each agent still acts based solely on local information. MADDPG \cite{NIPS2017_68a97503} is a well-known example of this approach and has inspired many variations applied to freshness optimization, including UAV coordination in \cite{9322539} and \cite{10293964}. Communication-based MARL methods also help reduce uncertainty by enabling agents to share limited information about their intentions or actions, thereby improving coordination. Opponent modeling techniques, such as LOLA \cite{10.5555/3237383.3237408}, go a step further by predicting how other agents will modify their strategies, making it easier to learn stable policies in competitive or mixed environments.

Recent work has applied these ideas to information freshness more directly. In \cite{10077432}, a modified version of MADDPG is proposed for vehicular platoons. The method introduces a global critic for team-level coordination and individual critics for each vehicle, which improves privacy and reduces the amount of shared experience required during training. This hierarchical design helps stabilize learning and makes it more scalable for large networks.

%Partial observability and non-stationarity, therefore, represent two fundamental obstacles to the application of MARL in freshness-driven communication systems. Understanding and addressing these difficulties is essential for designing agents that can operate reliably in large wireless networks where actions are tightly coupled, and timing is critical for maintaining information freshness.

Partial observability and non-stationarity are intrinsic characteristics of MARL in freshness-driven communication systems. They influence how agents form beliefs about the network state and adapt their policies over time, with direct implications for reliable operation in large wireless networks where actions are tightly coupled and timing is critical for information freshness.

\section{Challenges and Future Research Directions}
\label{future}
This section outlines future research directions, identifies key challenges and opportunities in RL-based AoI optimization, and highlights potential areas for further exploration.

\subsection{Delayed MDP} 
In many aforementioned RL applications for AoI optimization, the reward design is directly tied to AoI-related performance metrics. Typically, the reward is structured around the reduction of AoI, encouraging the agent to learn strategies that keep information as fresh as possible. However, in real-world scenarios, the agent often receives delayed feedback, rendering the original MDP a POMDP \cite{9069178}. For example, in CSMA protocols, the agent learns the AoI only after receiving an ACK from the destination, meaning the reward is delayed.

This delayed feedback creates challenges for the learning process. Since the agent does not receive timely and informative feedback, it struggles to learn effective policies, particularly in environments where actions taken may not yield immediate observable results. This delayed feedback structure can lead to slower convergence and suboptimal policy learning because the agent might not receive sufficient guidance on whether its actions are leading to desirable outcomes \cite{NEURIPS2019_54e36c5f}. Furthermore, based on the natural feature of the environment, there could be either a constant delay or a stochastic delay, meaning the event of a delay is known/unknown \cite{9069178}, and there could also be the state delay, action delay, or reward delay \cite{1193736}.

To address the random delay issue, one potential approach in the domain of RL is the concept of partial trajectory resampling proposed by \cite{bouteiller2021reinforcement}. Given that actions in a delayed environment may not receive immediate feedback, this technique resamples parts of the trajectory to simulate what would have happened if the actions were taken under the current policy. By doing so, the agent can more accurately estimate the value of delayed actions, reducing the bias and variance in the learning process. 

Another potential approach is belief-based learning, which estimates system states when feedback is uncertain or unavailable. Belief learning, as proposed in \cite{10622227}, introduces a belief distribution that tracks the likelihood of different system states instead of assuming perfect state observation. At each time step, this belief is updated using available feedback, adjusting probabilities based on received or missing ACK/NACK messages. If an ACK is received, the belief becomes concentrated, meaning the system state is known with certainty. If feedback is lost, the belief remains distributed, reflecting uncertainty. This method allows the agent to make decisions based on estimated rather than fully observed states, effectively addressing delays in feedback reception. By incorporating belief-based updates into RL, the agent can infer hidden AoI states even in environments with unreliable feedback, such as CSMA protocols where ACK reception is inconsistent. This approach improves learning efficiency and policy performance, making it a viable approach for AoI minimization in wireless networks.

Another promising direction is world-model-based RL, which enables agents to learn predictive representations of environment dynamics and estimate current or future states when immediate feedback is unavailable. Karamzade et al. \cite{karamzade2024reinforcement} investigated world-model-based RL under delayed observations, demonstrating how learned dynamics can support decision-making in partially observable environments. More recently, Wang et al. \cite{wang2025worldmodel} proposed a world-model-based learning framework for long-term AoI optimization in vehicular networks, leveraging imagined trajectories to improve data efficiency and freshness-aware decision-making. Building on these advances, world models could be applied to AoI optimization under delayed ACK/NACK feedback by predicting the evolution of information freshness from transmission histories, channel conditions, and estimated packet delivery outcomes. Such predictive capabilities may help agents make decisions before feedback becomes available and improve long-term policy learning. Nevertheless, their effectiveness depends on the accuracy of learned network dynamics, particularly under stochastic interference, unpredictable traffic, and time-varying channel conditions. Developing uncertainty-aware world models for freshness optimization under delayed feedback therefore represents a promising future research direction.

\subsection{Cross-Layer Design for freshness Optimization in Wireless Networks}
As the complexity and demands of wireless networks continue to grow, optimizing AoI across multiple layers of the network stack offers a significant opportunity to enhance overall network performance. Cross-layer design, which involves the joint optimization of different layers such as the physical, MAC, network, and application layers, holds great promise for achieving more holistic AoI optimization. By leveraging RL techniques in this cross-layer design, it is possible to dynamically adapt and optimize AoI in response to varying network conditions, leading to substantial improvements in information freshness across the network.

The importance of cross-layer design for AoI optimization lies in the interconnected nature of the various network layers. For example, decisions made at the physical layer, such as transmission power, directly impact the data rate, which in turn influences AoI. Similarly, MAC-layer access control mechanisms play a critical role in determining the efficiency and timeliness of data transmission \cite{10020660,9386228,9798137,Li2023LearningCW}, while network-layer routing decisions determine the path and delay experienced by packets \cite{9316802}. Even at the application layer, factors such as sampling rates and scheduling are critical to maintaining low AoI, as discussed in \ref{sampling} and \ref{scheduling}. By considering these layers collectively rather than in isolation, cross-layer design can ensure that optimization at one layer does not inadvertently compromise the performance at another, leading to a more balanced and effective overall network operation.

However, cross-layer design is inherently complex due to correlations across layers. Decisions made at one layer can affect other layers, making it challenging to achieve a globally optimal solution.

To this end, hierarchical RL (HRL)\cite{Kulkarni2016HierarchicalDR} emerges as a promising solution to address these challenges. HRL structures the learning process into multiple levels, where higher-level policies make more abstract decisions and lower-level policies handle more detailed, layer-specific tasks. This hierarchical approach can significantly reduce the complexity of cross-layer optimization by breaking it down into manageable sub-tasks.

For example, the work in \cite{10083085} adopts a cross-layer design to optimize AoI by jointly considering the physical and network layers. At the physical layer, the AP’s beamforming strategy and the IRS’s phase-shifting matrix are optimized. At the network layer, user transmissions are scheduled. These components are coordinated within a hierarchical RL framework, effectively addressing the AoI optimization challenge holistically.
    
\subsection{Against Randomness} 
In many real-world applications of RL for wireless networks, the environment is inherently stochastic (e.g., random noise and interference), leading to significant variability in rewards and action outcomes. Traditional RL methods typically aim to maximize the expected return, which is the average of all possible outcomes. However, this approach can be limiting in environments such as wireless networks with high variability or uncertainty, as it does not provide insight into the distribution of possible returns. For example, if two actions have the same expected return but differ in variance, the distribution of possible outcomes will influence the agent's decision based on their risk preferences. Therefore, understanding not only the expected return but also the full distribution of possible outcomes can offer a more nuanced policy.

Distributional RL\cite{10.5555/3305381.3305428} is an emerging approach that addresses this challenge of environmental randomness by estimating the entire distribution of returns rather than just their expectation. Instead of learning a single value representing the expected return, distributional RL algorithms learn a probability distribution over possible returns for each action. This distributional perspective allows the agent to capture the variability and uncertainty in outcomes more effectively.

In \cite{8931561}, the authors use a distributional RL framework, which is used against the randomness of the network to eliminate the effect of underestimating/overestimating the expected value of return in traditional RL to optimize the allocation of bandwidth among different network slices to ensure both spectral efficiency and service level agreement satisfaction ratio are maximized. 

However, the application of distributional RL to minimize AoI has not been thoroughly explored. While distributional RL has shown great potential in other domains by capturing the full distribution of return, its application in AoI optimization could also bring some significant advantages, like investigating the risk associated policy with extreme AoI values\cite{inproceedings} in network environments with variable delays, dynamic traffic or communication conditions.

\section{Conclusion}
\label{conclusion}
This survey presented an RL-centered perspective on information freshness in wireless networks. By organizing AoI into native, function-based, and application-oriented metrics, we highlighted how freshness objectives have evolved beyond simple recency to incorporate risk awareness and task relevance. We further introduced a policy-centric taxonomy that connects RL methods to key freshness-related decisions, including update control, medium access, risk-sensitive optimization, and multi-agent coordination. The surveyed studies demonstrate the potential of RL to learn adaptive policies under complex and uncertain network dynamics without requiring precise system models. Nevertheless, challenges related to delayed feedback, partial observability, and scalable multi-agent coordination remain open. Addressing these challenges is essential for advancing practical RL-based freshness optimization in B5G and 6G wireless networks. Overall, this survey provides a unified foundation for understanding existing approaches and guiding future research toward adaptive and application-aware freshness optimization.

\section*{Acknowledgment}
This work was supported in part by the NSF under
grant CNS-2008092.

\bibliographystyle{unsrtnat}
\bibliography{references}  %%% Uncomment this line and comment out the ``thebibliography'' section below to use the external .bib file (using bibtex) .

%%% Uncomment this section and comment out the \bibliography{references} line above to use inline references.
% \begin{thebibliography}{1}

% 	\bibitem{kour2014real}
% 	George Kour and Raid Saabne.
% 	\newblock Real-time segmentation of on-line handwritten arabic script.
% 	\newblock In {\em Frontiers in Handwriting Recognition (ICFHR), 2014 14th
% 			International Conference on}, pages 417--422. IEEE, 2014.

% 	\bibitem{kour2014fast}
% 	George Kour and Raid Saabne.
% 	\newblock Fast classification of handwritten on-line arabic characters.
% 	\newblock In {\em Soft Computing and Pattern Recognition (SoCPaR), 2014 6th
% 			International Conference of}, pages 312--318. IEEE, 2014.

% 	\bibitem{hadash2018estimate}
% 	Guy Hadash, Einat Kermany, Boaz Carmeli, Ofer Lavi, George Kour, and Alon
% 	Jacovi.
% 	\newblock Estimate and replace: A novel approach to integrating deep neural
% 	networks with existing applications.
% 	\newblock {\em arXiv preprint arXiv:1804.09028}, 2018.

% \end{thebibliography}

\end{document}